\documentclass[lettersize,journal]{IEEEtran}
\usepackage{amsmath,amsfonts}
\usepackage{algorithmic}
\usepackage{algorithm}
\usepackage{array}
\usepackage[caption=false]{subfig}
\usepackage{textcomp}
\usepackage{stfloats}
\usepackage{url}
\usepackage{verbatim}
\usepackage{graphicx}
\usepackage{cite}
\usepackage{amsmath}
\usepackage{amssymb}
\usepackage{multirow}
\usepackage{bbding}
\usepackage{setspace}
\usepackage{threeparttable}
\usepackage{balance}
\usepackage{tikz,xcolor}

\usepackage[colorlinks,
linkcolor=blue,
anchorcolor=blue,
citecolor=blue]{hyperref}

\hypersetup{hidelinks,
	colorlinks=true,
	allcolors=blue,
	pdfstartview=Fit,
	breaklinks=true}

\definecolor{lime}{HTML}{A6CE39}
\DeclareRobustCommand{\orcidicon}{
\begin{tikzpicture}
\draw[lime, fill=lime] (0,0)
circle[radius=0.16]
node[white]{{\fontfamily{qag}\selectfont \tiny \.{I}D}};
\end{tikzpicture}
\hspace{-2mm}
}
\foreach \x in {A, ..., Z}{%
\expandafter\xdef\csname orcid\x\endcsname{\noexpand\href{https://orcid.org/\csname orcidauthor\x\endcsname}{\noexpand\orcidicon}}
}

\begin{document}

\title{CFN-ESA: A Cross-Modal Fusion Network With Emotion-Shift Awareness for Dialogue Emotion Recognition}
\author{Jiang Li\hspace{-1.5mm}\orcidA{}, Xiaoping Wang\hspace{-1.5mm}\orcidB{},~\IEEEmembership{Senior Member,~IEEE}, Yingjian Liu, and Zhigang Zeng\hspace{-1.5mm}\orcidC{},~\IEEEmembership{Fellow,~IEEE}
        
\thanks{Manuscript received 7 September 2023; revised 5 February 2024; accepted 11 April 2024. This work was supported in part by the National Natural Science Foundation of China under Grant 62236005 and 61936004. \textit{(Corresponding author: Xiaoping Wang.)}}
\thanks{The authors are with the School of Artificial Intelligence and Automation, Huazhong University of Science and Technology (HUST), the Institute of Artificial Intelligence, HUST, the Hubei Key Laboratory of Brain-inspired Intelligent Systems, HUST, and the Key Laboratory of Image Processing and Intelligent Control (HUST), Ministry of Education, Wuhan 430074, China (e-mail: lijfrank@hust.edu.cn; wangxiaoping@hust.edu.cn; virtualmoon999@gmail.com; zgzeng@hust.edu.cn).}
\thanks{Digital Object Identifier 10.1109/TAFFC.2024.3389453}
}

\markboth{IEEE TRANSACTIONS ON AFFECTIVE COMPUTING}
{LI \MakeLowercase{\textit{et al.}}: A Cross-Modal Fusion Network With Emotion-Shift Awareness for Dialogue Emotion Recognition}

\IEEEpubid{0000--0000~\copyright~2024 IEEE. Personal use is permitted, but republication/redistribution requires IEEE permission.}

\maketitle

\begin{abstract}
Multimodal emotion recognition in conversation (ERC) has garnered growing attention from research communities in various fields. In this paper, we propose a Cross-modal Fusion Network with Emotion-Shift Awareness (CFN-ESA) for ERC. Extant approaches employ each modality equally without distinguishing the amount of emotional information in these modalities, rendering it hard to adequately extract complementary information from multimodal data. To cope with this problem, in CFN-ESA, we treat textual modality as the primary source of emotional information, while visual and acoustic modalities are taken as the secondary sources. Besides, most multimodal ERC models ignore emotion-shift information and overfocus on contextual information, leading to the failure of emotion recognition under emotion-shift scenario. We elaborate an emotion-shift module to address this challenge. CFN-ESA mainly consists of unimodal encoder (RUME), cross-modal encoder (ACME), and emotion-shift module (LESM). RUME is applied to extract conversation-level contextual emotional cues while pulling together data distributions between modalities; ACME is utilized to perform multimodal interaction centered on textual modality; LESM is used to model emotion shift and capture emotion-shift information, thereby guiding the learning of the main task. Experimental results demonstrate that CFN-ESA can effectively promote performance for ERC and remarkably outperform state-of-the-art models.
\end{abstract}


\section{Introduction}
\IEEEPARstart{R}{ecently}, multimodal learning has attracted the attention of both academia and industry, and has been widely applied in many fields, such as biometrics, information retrieval, autonomous driving, and emotion recognition. With the advancement of technologies, the abundance of multimodal data can be more conveniently available for research purposes. In realistic life, multimodal data mainly contains three contents, i.e., transcribed text, visual image or video, and acoustic speech. Multimodal fusion is one of the prominent branches of multimodal learning, whose main purpose is to utilize the organic combination of information from multiple modalities to collaboratively achieve the final downstream task. Thus, how to adequately extract the inter-modal complementary information becomes a formidable challenge in the domain of multimodal fusion.

\IEEEpubidadjcol
\begin{figure}[htbp]
    \centering
    \includegraphics[width=3.4in]{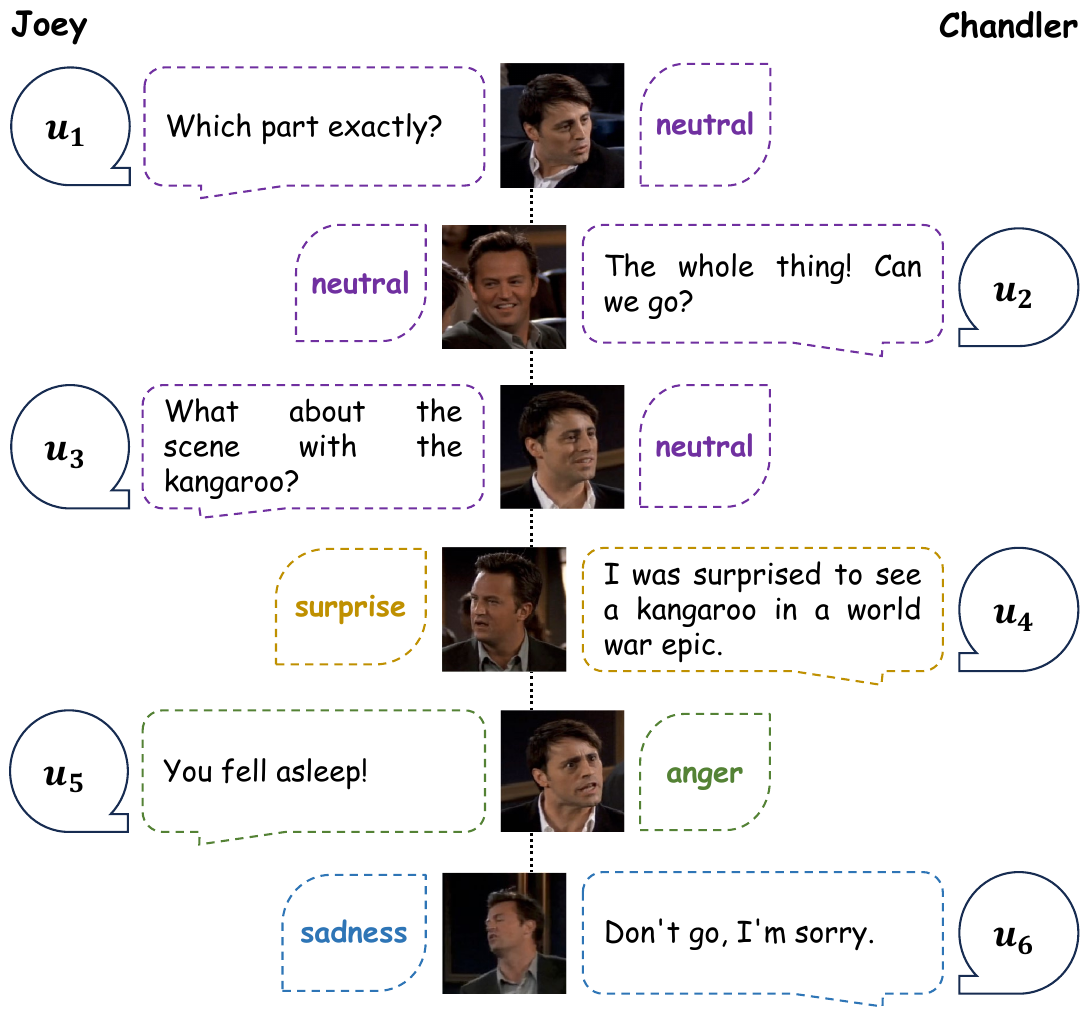}
    \caption{A conversational scene from the MELD dataset. If only textual modality is taken into account, the emotion of $u_5$ may be recognized as \textit{neutral}. From the facial expression of the speaker who utters $u_5$, it is known that the emotion should be \textit{anger}, which is true emotion of the utterance.}
    \label{fig:example}
\end{figure}
The target of emotion recognition in conversation (ERC) is to understand and analyze each utterance in the conversation and render the corresponding emotion. This task has recently drawn widespread interest from researchers in the areas of natural language processing, computer vision, and multimodal learning due to its promising applications, such as human-machine interface in intelligent robots and opinion mining in social media. Most previous ERC models are based on individual modalities, such as text~\cite{jiao2020real,shen2021dialogxl,nie2022igcn,zhao2022cauain,li2022contrast} and speech~\cite{fan2022isnet,latif2022multitask,lei2022bat,zhou2022multiclassifier}. However, very often, the emotions of human beings are elusive. As shown in Fig.~\ref{fig:example}, textual uni-modality may not be capable of correctly recognizing emotions in some scenarios, e.g., the emotion directly expressed by the text is \textit{neutral}, but the corresponding facial expression is actually another emotion, e.g., \textit{anger}. From this example, it can be argued that the model cannot understand and convey human emotions well with only a single modality. As multi-modality gets closer to real-world application scenarios, multimodal ERC has gained numerous research. The information contained in a single modality may not be sufficient or representative enough, while a multimodal-based model can make up for the shortcoming of the unimodal approach and thus improve the performance and robustness of the existing system. Simultaneously, multimodal ERC is more in line with the multiple ways (e.g., language, voice, and facial expressions) in which people express their emotions. Unlike traditional affective computing missions in unimodal~\cite{jiao2020real,shen2021dialogxl,li2022contrast} and non-conversational~\cite{he2023multimodal,mai2022hybrid,yu2021learning} scenarios, multimodal ERC suffers from harsher challenges due to the complex relationship between multiple modalities and conversational contexts.

Although previous studies have made impressive progress, these approaches either ignore the association between multimodal information or model multi-modality insufficiently. Some methods~\cite{poria2017context,hazarika2018conversational,hazarika2018icon,majumder2019dialoguernn} directly concatenate multimodal data without considering the association between multiple modalities. Moreover, there is a certain amount of noise in each modality itself, and together with the heterogeneity gap~\cite{hazarika2020misa} of multimodal data, this direct concatenation manner may cause more noise. While some approaches~\cite{hu2021mmgcn,chen2021learning,hu2022mmdfn,chen2022modeling} perform associative modeling for multimodal data, there are flaws in their modeling styles. For instance, these methods assume that each modality contributes equally to the emotional expression of the utterance, which is not the case. The findings of extant multimodal ERC studies~\cite{mao2021dialoguetrm,hu2022mmdfn} indicate that textual modalities contain more valuable emotional information in comparison to visual and acoustic modalities. Consequently, exploiting each modality equally may not adequately extract multimodal complementary information when engaging in multimodal interaction, making it difficult to effectively maximize the performance of the model. Towards the above issues, we construct a novel network for conversational emotion recognition to efficiently model the association with multimodal data. We treat visual and acoustic modalities as sources of auxiliary information that are utilized to complement the representation of textual information; in turn, textual information is employed to augment visual and acoustic representations. 

Extant efforts~\cite{majumder2019dialoguernn,ghosal2020cosmic,shen2021directed} have revealed that emotion shift can constrain the performance of emotion recognition and is one of the challenges faced by ERC. Emotion shift describes the change of emotions in two utterances. More concretely, if two utterances shift from one emotion to another, i.e., the emotions of two utterances are different, then the emotion shift has occurred; conversely, the emotion shift has not occurred if the emotions of two utterances are identical. Contextual modeling, which inherently relies on aggregating emotional cues from surrounding utterances, often tends to preserve emotional consistency across the conversation. Nevertheless, this inherent tendency may inadvertently undermine the model's capacity to accurately recognize emotions under situations where emotion shifts occur, thus highlighting the need for advanced strategies to address this critical aspect of ERC. Existing approaches fail to consider emotion-shift information and concentrate too much on contextual information, causing the imbalance between context- and self-modeling. In other words, the importance of self-information (complementary information from the current utterance but belonging to the other two modalities) is prone to be neglected. To alleviate this problem, we devise an emotion-shift module as the auxiliary task of ERC, which guides the main task of ERC to optimize the emotional expression of utterances by taking into account emotion-shift factor.

To summarize, we propose a \textbf{C}ross-modal \textbf{F}usion \textbf{N}etwork with \textbf{E}motion-\textbf{S}hift \textbf{A}wareness (CFN-ESA) for ERC. Our CFN-ESA can efficiently extract multimodal complementary information, which mainly consists of three components, i.e., recurrence based uni-modality encoder (RUME), attention based cross-modality encoder (ACME), and label based emotion-shift module (LESM). RUME can capture intra-modal contextual emotional cues while narrowing the heterogeneity gap of multimodal data by sharing parameters. ACME perceives textual modality as the primary source of emotional information and two other modalities as the secondary sources, and employs multi-head attention networks to adequately model multimodal interaction. LESM is employed as an auxiliary task of the ERC to explicitly model emotion shift and extract emotion-shift information, thereby enabling the main task to implicitly reduce intra-modal contextual modeling under emotion-shift scenario. Two public benchmark datasets, MELD and IEMOCAP, are leveraged to conduct numerous experiments for demonstrating the effectiveness of the proposed CFN-ESA. We also explore the impact under different network settings and test the performance of each component in CFN-ESA. To put it in a nutshell, the main contributions of this work include:
\begin{enumerate}
	\item A novel multimodal ERC method named CFN-ESA is proposed, which is mainly composed of uni-modality encoder (RUME), cross-modality encoder (ACME), and emotion-shift module (LESM).
	\item RUME can extracts intra-modal contextual information while mitigating the heterogeneity gap issue; ACME can model multimodal interaction and adequately captures inter-modal complementary information.
	\item LESM is utilized as an auxiliary task of the model to extract emotion-shift information, which in turn guides the main task for learning.
	\item We conduct abundant experiments on two datasets and the results attest to the superiority of CFN-ESA over all baselines.
\end{enumerate}

\section{Related Works}\label{work}
\subsection{Emotion Recognition in Conversation}
With the mounting interest in the study of dialogue systems, the identification of emotion in the conversation has become a hot research topic. Most previous ERC methods are based on textual modality, which primarily employ gated recurrent unit (GRU), long and short term memory (LSTM) network, and graph neural network (GNN) to model contexts. 
AGHMN~\cite{jiao2020real} mainly consisted of hierarchical memory network (HMN) and bidirectional gated recurrent unit (BiGRU), where HMN was used to extract the interactive information between historical utterances, and BiGRU was used for the summarization of short- and long-term memory with the help of attentional weights. 
DialogXL~\cite{shen2021dialogxl} applied the pre-trained language model XLNet~\cite{yang2019xlnet} to the ERC task. To achieve this purpose, DialogXL handled long-term context with enhanced memory and speaker dependencies with dialogue-aware self-attention. 
I-GCN~\cite{nie2022igcn} utilized graph convolutional network to extract the semantic associative information of utterances and the temporal sequence information of dialogues. The method firstly exploited graph structure to represent dialogues at different times and then employed incremental graph structure to simulate the process of dynamic dialogues.
CauAIN~\cite{zhao2022cauain} consisted of two main causal-aware interactions, namely causal cue retrieval and causal utterance traceback, which introduced commonsense knowledge as a cue for detecting emotional causes in a dialogue, explicitly modeling intra- and inter-speaker dependencies.
CoG-BART~\cite{li2022contrast} was an ERC approach that employed both contrastive learning and generative modeling, which utilized BART~\cite{lewis2020bart} as a backbone model, and enhanced the emotional expression of utterances through contrastive loss and generative loss.

The approaches based on acoustic modality are often termed as speech emotion recognition (SER). ISNet~\cite{fan2022isnet} was an individual standardization network that adopted automatically generated benchmark for individual standardization to deal with the problem of inter-individual emotion confusion in SER. MTL-AUG~\cite{latif2022multitask} was a semi-supervised multitask learning framework that employed speech-based augmentation types, while treating augmented classification and unsupervised reconstruction as auxiliary tasks to enable multi-task training to achieve the learning of generic representations without the need for meta-labeling. BAT~\cite{lei2022bat} split the hybrid spectrogram into blocks and computed self-attention by combining these blocks with tokens, meanwhile utilizing the cross-block attention mechanism to facilitate the information interaction between blocks. In order to gain a deeper understanding of emotions conveyed in speech, Huang et al.~\cite{huang2015aninvestigation,huang2016detecting} carried out in-depth investigations on emotion change detection. These studies provided insights into emotion change and could inspire future work in the field of ERC. Furthermore, while there exist some visual modality-based methods~\cite{xu2022mdan,zhu2017dependency,she2020wscnet} known as facial expression recognition, they are mostly outside the scope of the ERC task.

There have been some multimodal ERC efforts recently. 
MMGCN~\cite{hu2021mmgcn} exploited GNN to capture contextual and modal interactive information, which not only compensated for the shortcomings of previous methods that are unable to leverage multimodal dependencies, but also efficiently incorporated the speaker's information for ERC. 
DialogueTRM~\cite{mao2021dialoguetrm} used hierarchical Transformer to manage differentiated contextual preferences within each modality, and designed multi-grained interactive fusion to learn the different contributions of multiple modalities. 
MetaDrop~\cite{chen2021learning} presented a dyadic contain or drop decision-making mechanism to learn adaptive fusion paths while extracting multimodal dependencies and contextual relationships.
HU-Dialogue~\cite{chen2022modeling} introduced hierarchical uncertainty for ERC, containing a regularization based attention module that was perturbed by source-adaptive noise to model context-level uncertainty. 
MM-DFN~\cite{hu2022mmdfn} utilized a graph based dynamic fusion module to track conversational contexts in various semantic spaces and to enhance complementarity between modalities.
COGMEN~\cite{joshi2022cogmen} was a multimodal ERC model that used a GNN architecture to model local dependencies and global contexts in the conversation, which effectively improved the performance of the model. 
UniMSE~\cite{hu2022unimse} integrated acoustic and visual features with textual features by applying T5~\cite{raffel2020exploring}, and performed inter-modal contrastive learning to obtain differentiated multimodal representations. 
Inspired by the phenomenon of emotional ups and downs in conversations, Agarwal et al.~\cite{bansal-etal-2022-shapes} proposed an emotion-shift component to enhance the performance of multimodal ERC. We observe that their presented method aligns with a similar research trajectory to the method in our paper. 
In general, distinct from traditional affective computing tasks in single-modal and non-conversational settings, multimodal ERC is more challenging due to the complex relationship of multiple modalities and dialogue contexts. 

\subsection{Multi-Head Attention Network}
Vaswani et al.~\cite{vaswani2017attention} proposed the Transformer architecture for machine translation task, which achieved exceptional performance. Since then, the multi-head attention (MHA) network of Transformer has been widely applied in the fields of natural language processing, computer vision, and multimodal learning. MulT~\cite{tsai-etal-2019-multimodal} employed multiple attention networks to model interactions among multimodal sequences with varying temporal steps, serving the purpose of multimodal sentiment analysis. AuxFormer~\cite{Goncalves2022AuxFormer} utilized a main audio-visual fusion network based on multi-head attention to achieve multimodal alignment and fusion, while two auxiliary networks were used to make the emotion information flow to the main network. Wagner et al.~\cite{Wagner2023Dawn} revealed through extensive experiments that Transformer-based speech emotion recognition exhibited higher robustness and generalizability relative to other architecture-based approaches. ViT~\cite{dosovitskiy2021an} applied pure Transformer directly to image sequence patches and achieved superior outcomes with few computational resources. BLIP-2~\cite{Li2023BLIP2} guided vision-language pre-training from frozen pre-trained image encoders and frozen large language models, and compensated for the modality gap with a lightweight query Transformer. LLaVA~\cite{liu2023visual} achieved universal vision-language understanding by bridging a visual encoder and a large language model, and facilitated future studies on visual instruction following. In this paper, we adopt MHA networks to extract multimodal complementary information, i.e., they are utilized to construct attention based cross-modality encoder (ACME). Here, the scaled dot-product attention is first defined:
\begin{equation}
    \label{eq:attention}
    \mathrm{ATT}(\mathrm{Q},\mathrm{K},\mathrm{V})=\mathtt{SMAX}\left (\frac{\mathrm{Q}\cdot \mathrm{K}^\top}{\sqrt{d_k}}\right )\cdot \mathrm{V},
\end{equation}
where query $\mathrm{Q}$, key $\mathrm{K}$, and value $\mathrm{V}$ are the packed feature representations; $d_k$ denotes the dimension of $\mathrm{K}$ or $\mathrm{V}$; $\mathtt{SMAX}(\cdot)$ denotes the softmax function. MHA is a network structure that can enhance the stability and performance of the scaled dot-product attention. The distinction is that different heads employ different query, key, and value matrices. MHA can be computed as follows:
\begin{equation}
    \label{eq:multi_attention}
    \begin{split}
    &\mathtt{MHA}(\mathrm{Q},\mathrm{K},\mathrm{V})={W}_{mha}\mathtt{CAT}(\mathrm{head}_0, \cdots, \mathrm{head}_h),\\
    &\mathbf{s.t.}\ \mathrm{head}_i=\mathtt{ATT}(W_{\mathrm{Q},i} \mathrm{Q}, W_{\mathrm{K},i} \mathrm{K},W_{\mathrm{V},i} \mathrm{V}),
    \end{split}
\end{equation}
where $\mathtt{CAT}(\cdot)$ denotes the concatenation operation; $W_{\mathrm{Q},i}$, $W_{\mathrm{K},i}$, and $W_{\mathrm{V},i}$ are the learnable parameters, which can project $\mathrm{Q}$, $\mathrm{K}$, and $\mathrm{V}$ into different representation subspaces, respectively; $\mathrm{W}_{mha}$ is also the trainable parameter.

\section{Proposed Model}\label{model}
\begin{figure*}[htbp]
    \centering
    \includegraphics[width=7.1in]{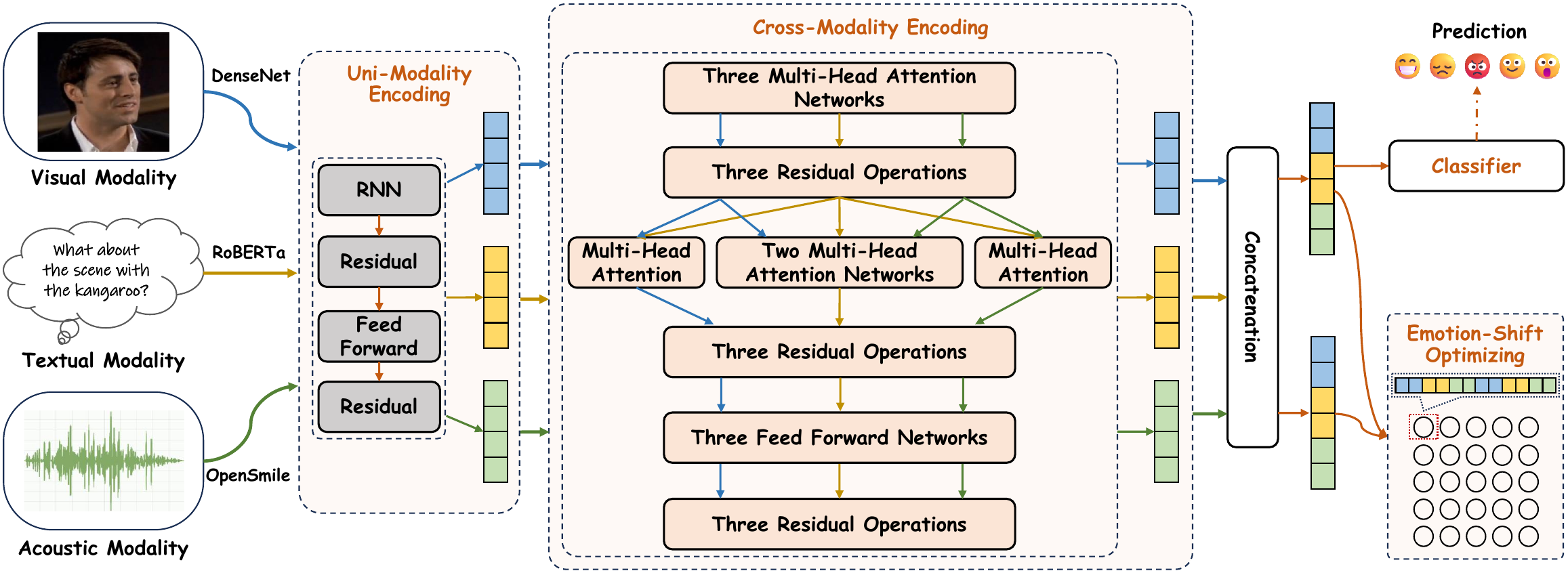}
    \caption{The overall architecture of our CFN-ESA. First, the utterance-level features of visual, textual, and acoustic modalities are extracted by DenseNet, RoBERTa, and OpenSmile, respectively; second, the intra-modal contextual information and inter-modal complementary information are captured by uni-modality encoder and cross-modality encoder in turn; then, the optimization of the utterance expression is performed by utilizing the emotion-shift module; finally, the emotion classifier is adopted for prediction.}
    \label{fig:overall_architecture}
\end{figure*}
This section is a detailed description of our proposed model. As shown in Fig.~\ref{fig:overall_architecture}, CFN-ESA mainly consists of the recurrence based uni-modality encoder (Uni-Modality Encoding), attention based cross-modality encoder (Cross-Modality Encoding), emotion classifier (Classifier), and label based emotion-shift module (Emotion-Shift Optimizing).

\subsection{Problem Definition}
Given a conversation $U$ containing $|U|$ utterances $u_1$, $\cdots$, $u_{|U|}$, i.e., $U = \{u_1,\cdots,u_{|U|}\}$, the goal of ERC is to predict the emotion state $e_i$ for each utterance $u_i$ in $U$. In other words, the task of ERC is to learn a function $\mathtt{F}(\cdot)$ with learnable parameters that maps the feature representation $x_i$ of an utterance $u_i$ to the corresponding emotion $e_i$, i.e., $e_i = \mathtt{F}(x_i)$. Here, a conversation is expressed by $|M|$ different modalities, i.e., $U=\{U_{m_1},\cdots,U_{m_{|M|}}\}$; and the set of modalities can be represented as $M=\{m_1,\cdots,m_{|M|}\}$. In our work, a conversation involves three modalities, i.e., textual ($T$), visual ($V$), and acoustic ($A$) modalities, so each utterance $u_i$ can be represented as $u_i=\{u_i^T,u_i^V,u_i^A\}$.

\subsection{Recurrence Based Uni-Modality Encoder}
\begin{figure}[htbp]
    \centering
    \includegraphics[width=3.4in]{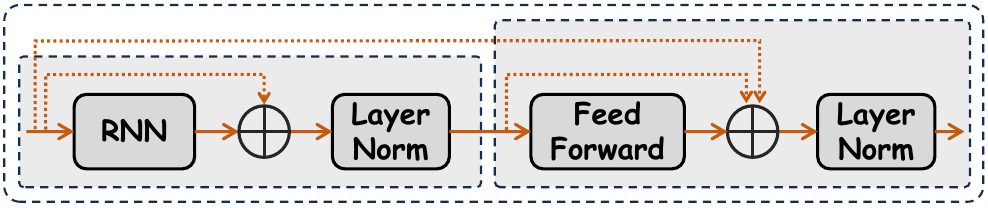}
    \caption{The network structure of RUME. Note that RUME shares parameters for each modality, and $\oplus$ denotes the residual operation.}
    \label{fig:rume}
\end{figure}
To extract dialogue-level contextual emotional cues, we employ recurrence based uni-modality encoder (RUME) to encode the utterance in each of three modalities. Inspired by the structure of Transformer~\cite{vaswani2017attention}, we add fully connected networks and residual operations to RUME to improve the expressiveness and stability of recurrent neural network (RNN). Our uni-modality encoder is shown in Fig.~\ref{fig:rume}. Specifically, the structure of RUME can be formalized as:
\begin{equation}
    \begin{split}
        &X_{rr}=\mathtt{LN}(X + \mathtt{RNN}(X)),\\
        &X_{fr} = \mathtt{LN}(X + X_{rr} + \mathtt{FF}(X_{rr})),
    \end{split}
\end{equation}
where $X$ denotes the feature matrix of all utterances; $\mathtt{RNN}(\cdot)$, $\mathtt{LN}(\cdot)$, and $\mathtt{FF}(\cdot)$ denote the RNN, normalization, and feedforward network layers, respectively. In this work, the $\mathtt{RNN}(\cdot)$ and $\mathtt{LN}(\cdot)$ default to bidirectional GRU and layer normalization; while the feed-forward layer consists of two fully connected networks, which can be represented as,
\begin{equation}
	\mathtt{FF}(X_{rr}) = \mathtt{DP}(\mathtt{FC}(\mathtt{DP}(\alpha(\mathtt{FC}(X_{rr}))))),
\end{equation}
where $\mathtt{FC}(\cdot)$ and $\mathtt{DP}(\cdot)$ denote the fully connected network and dropout operation, respectively, and $\alpha(\cdot)$ denotes the activation function.

Note that in order to make the data distribution for each modal utterance as close as possible (i.e., to alleviate the heterogeneity gap problem for multimodal data), we utilize the uni-modality encoder with shared parameter for all three modalities. That is, $X_{fr}^{m} = \mathtt{RUME}(X^{m})$, where $m \in \{T,V,A\}$ and $\mathtt{RUME}(\cdot)$ denotes the uni-modality encoder.

\subsection{Attention Based Cross-Modality Encoder}
\begin{figure*}[htbp]
    \centering
	\subfloat[Visual Part]{\includegraphics[height=3.1in]{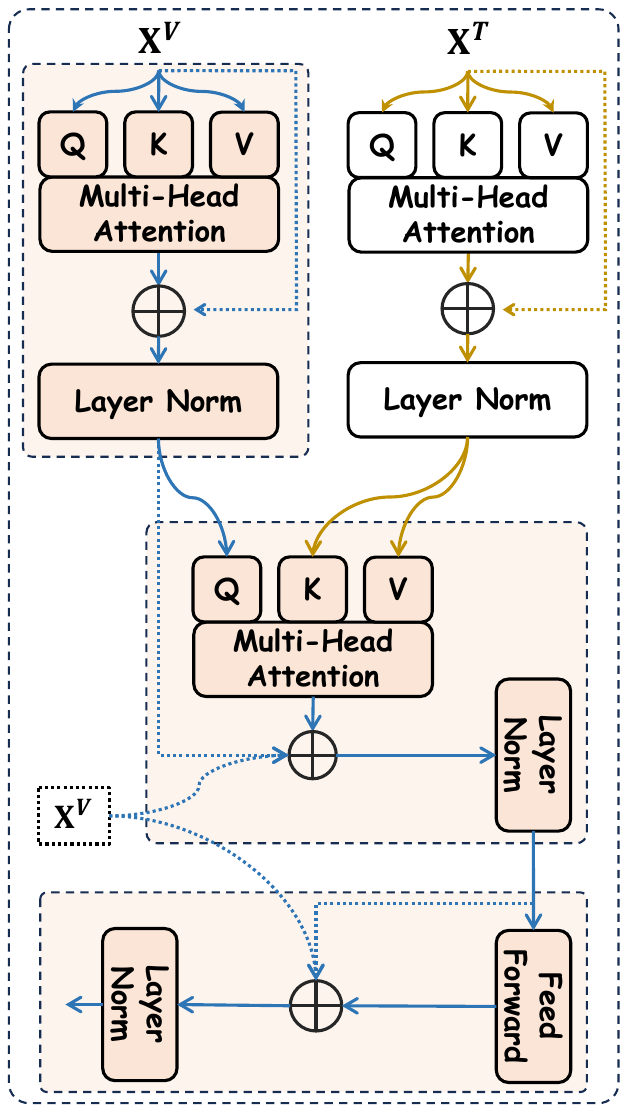}%
    \label{fig:acme_v}}
    \hfil
    \subfloat[Textual Part]{\includegraphics[height=3.1in]{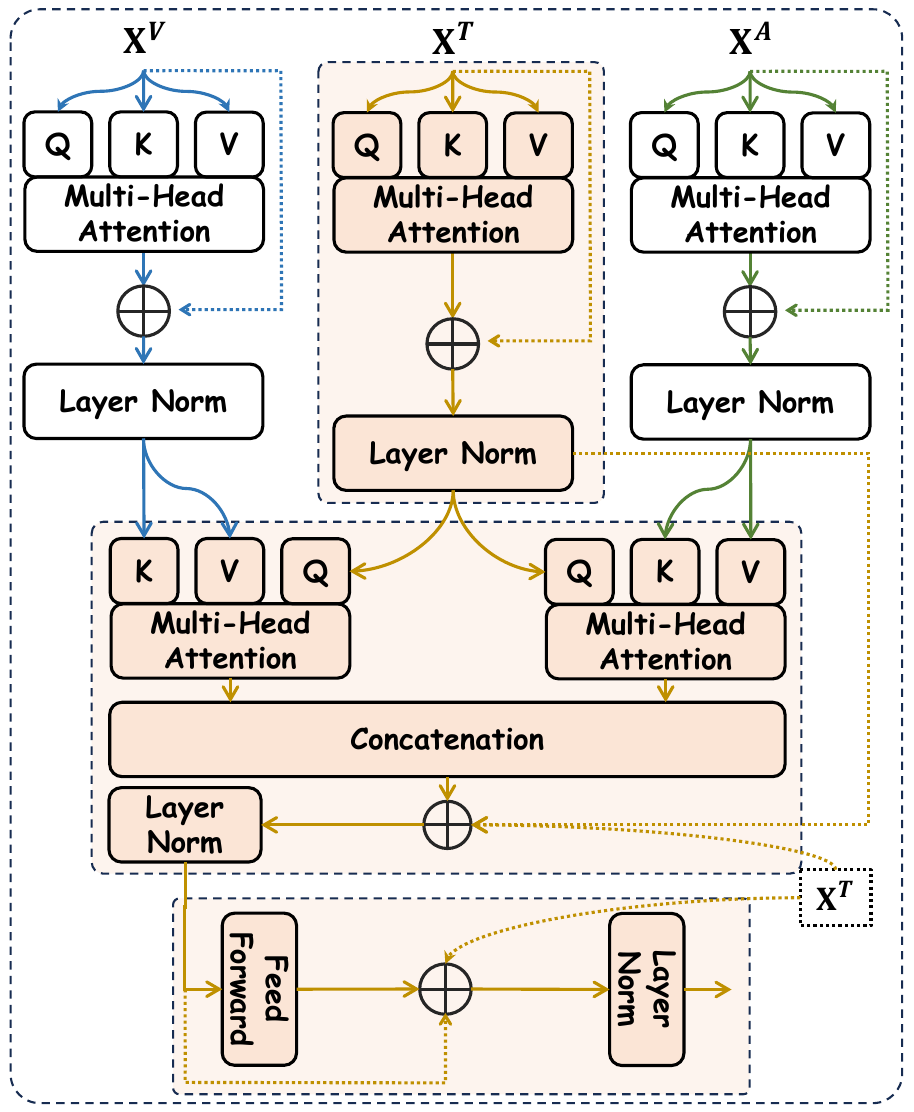}%
    \label{fig:acme_t}}
	\hfil
    \subfloat[Acoustic Part]{\includegraphics[height=3.1in]{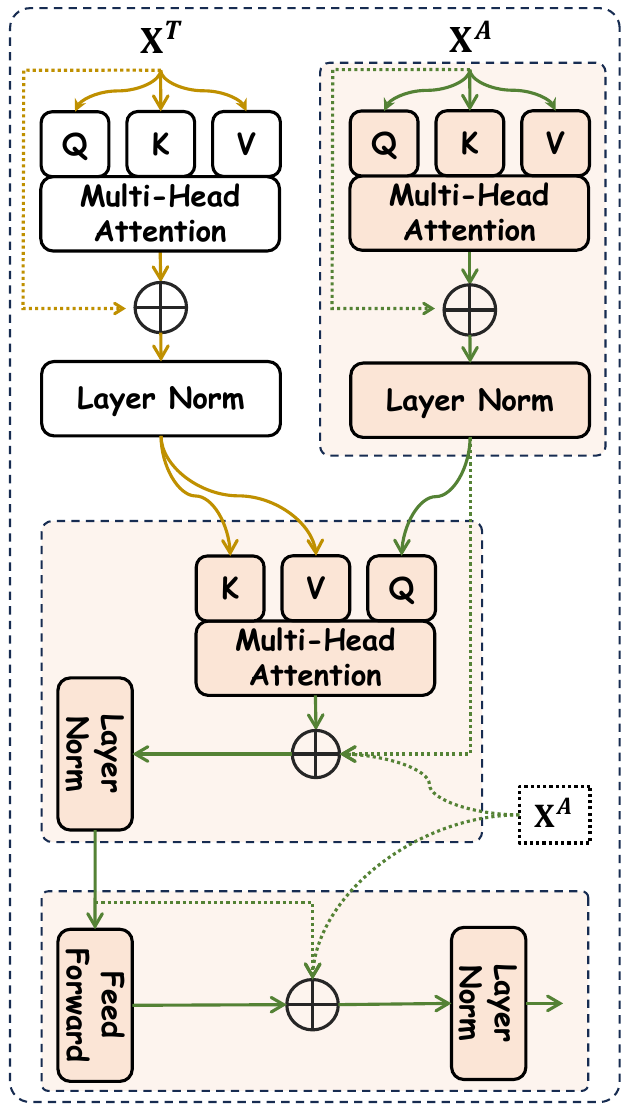}%
    \label{fig:acme_a}}
    \caption{The network structure of ACME. (a), (b), and (c) show the structure for visual, textual, and acoustic information updating in ACME, respectively. Note that the information updating network for visual modality is similar to that for acoustic modality.}
    \label{fig:acme}
\end{figure*}
Multimodal ERC can compensate for the lack of information in unimodal methods. In this work, we devise attention based cross-modality encoder (ACME) to extract complementary information from multimodal emotion data. As shown in Fig.~\ref{fig:acme}, we take inspiration from the Transformer structure and mainly adopt the attention network layer, feedforward network layer, and residual operation to construct our ACME. Several studies~\cite{hu2021mmgcn,mao2021dialoguetrm} on multimodal ERC have revealed that the amount of emotional information embedded in visual and acoustic modalities is lower than that in textual modalities, and thus the expression of emotion in these models is limited. Based on this assumption, we take both visual and acoustic features as complementary information to complement the emotional expression of textual features. In turn, textual features of utterances are used to enhance the visual and acoustic representations. Furthermore, in RUME, it is laborious for RNN to focus on the global contextual information of the utterance. Therefore, we employ a self-attention layer to capture global contextual emotional cues before performing cross-modal interaction. The designed ACME is composed of the following three stages.

(1) Enhancing the global contextual awareness of the utterance. The feature matrices $\mathrm{X}^m$ from three modalities are taken as the inputs to three MHA networks, and the direct output $\mathrm{X}_{s}^m$ is summed with the input $\mathrm{X}^m$ (i.e., the residual operation) to obtain feature matrix $\mathrm{X}_{sr}^m$. This process can be expressed by equations as:
\begin{equation}
    \begin{split}
        &\mathrm{X}_{s}^m = \mathtt{DP}(\mathtt{MHA}(\mathrm{X}^m,\mathrm{X}^m,\mathrm{X}^m)),\\
		&\mathrm{X}_{sr}^m = \mathtt{LN}(\mathrm{X}^m+\mathrm{X}_{s}^m),
    \end{split}
\end{equation}
where $\mathtt{MHA}(\cdot)$ denotes the MHA network.

(2) Performing the cross-modal interaction modeling. The above results are employed as inputs to four MHA networks in pairwise manner, and the information for each modality is updated. In the following, we describe the information update for each modality separately.

For the information update in textual modality, there are mainly two MHA networks and the feature matrices from three modalities being leveraged. Specifically, the textual feature matrix $\mathrm{X}_{sr}^T$ is utilized as the query $\mathrm{Q}$ in one MHA network, and the visual feature matrix $\mathrm{X}_{sr}^V$ is utilized as the key $\mathrm{K}$ and the value $\mathrm{V}$, and the output $\mathrm{X}_{c} ^{T \leftarrow V}$ is a textual feature matrix with visual information; similarly, the query $\mathrm{Q}$ in another MHA network comes from $\mathrm{X}_{sr}^T$, the key $\mathrm{K}$ and value $\mathrm{V}$ are $\mathrm{X}_{sr}^A$, and we obtain $\mathrm{X}_{c}^{T \leftarrow A}$, a textual feature matrix with acoustic information; we further concatenate $\mathrm{X}_{c}^{T \leftarrow V}$ and $\mathrm{X}_{c}^{T \leftarrow A}$ to get $\mathrm{X}_{c}^{T}$, and at the same time, we apply the residual operation to add $\mathrm{X}^T$, $\mathrm{X}_{sr}^T$, and $\mathrm{X}_{c}^{T}$ to obtain the new textual feature matrix $\mathrm{X}_{cr}^{T}$. The above process can be formalized as:
\begin{equation}
    \begin{split}
        &\mathrm{X}_{c}^{T \leftarrow V} = \mathtt{DP}(\mathtt{MHA}(\mathrm{X}_{sr}^T,\mathrm{X}_{sr}^V,\mathrm{X}_{sr}^V)),\\
		&\mathrm{X}_{c}^{T \leftarrow A} = \mathtt{DP}(\mathtt{MHA}(\mathrm{X}_{sr}^T,\mathrm{X}_{sr}^A,\mathrm{X}_{sr}^A)),\\
		&\mathrm{X}_{c}^{T} = \alpha(\mathtt{FC}(\mathtt{CAT}(\mathrm{X}_{c}^{T \leftarrow V},\mathrm{X}_{c}^{T\leftarrow A}))),\\
		&\mathrm{X}_{cr}^{T} = \mathtt{LN}(\mathrm{X}^T+\mathrm{X}_{sr}^T+\mathrm{X}_{c}^T),
    \end{split}
\end{equation}
where $\mathtt{CAT}(\cdot)$ represents the concatenation operation.

For the information update in visual modality, one attention network and the feature matrices from two modalities are mainly used. Specifically, we take the visual feature matrix $\mathrm{X}_{sr}^V$ as the query $\mathrm{Q}$ in the MHA network, and the textual feature matrix $\mathrm{X}_{sr}^T$ as the key $\mathrm{K}$ and value $\mathrm{V}$, to obtain the visual feature matrix $\mathrm{X}_{c}^{V\leftarrow T}$ with textual information enhancement; similar to the textual information updating process, the residual operation is applied to add $\mathrm{X}^V$, $\mathrm{X}_{sr}^V$, and $\mathrm{X}_{c}^{V\leftarrow T}$ to gain the new visual feature matrix $\mathrm{X}_{cr}^{V}$. The above process can be formalized as: 
\begin{equation}
    \begin{split}
        &\mathrm{X}_{c}^{V\leftarrow T} = \mathtt{DP}(\mathtt{MHA}(\mathrm{X}_{sr}^V, \mathrm{X}_{sr}^T, \mathrm{X}_{sr}^T)),\\
		&\mathrm{X}_{cr}^{V} = \mathtt{LN}(\mathrm{X}^V + \mathrm{X}_{sr}^V + \mathrm{X}_{c}^{V\leftarrow T}).
    \end{split}
\end{equation}

The information updating process in acoustic modality is similar to that in visual modality, which can be expressed by the following equation:
\begin{equation}
    \begin{split}
        &\mathrm{X}_{c}^{A\leftarrow T} = \mathtt{DP}(\mathtt{MHA}(\mathrm{X}_{sr}^A, \mathrm{X}_{sr}^T, \mathrm{X}_{sr}^T)),\\
		&\mathrm{X}_{cr}^{A} = \mathtt{LN}(\mathrm{X}^A + \mathrm{X}_{sr}^A + \mathrm{X}_{c}^{A\leftarrow T}).
    \end{split}
\end{equation}

(3) Improving the expressiveness and stability of the model. We take $\mathrm{X}_{cr}^m$ as the input to each of three feedforward network layers to obtain $\mathrm{X}_{f}^m$; at the same time, the residual operation is used to sum $\mathrm{X}^m$, $\mathrm{X}_{cr}^m$, $\mathrm{X}_{f}^m$ to obtain the feature matrix $\mathrm{X}_{fr}^m$. The above process is expressed by the equation as follows: 
\begin{equation}
    \begin{split}
        &\mathrm{X}_{f}^m = \mathtt{FF}(\mathrm{X}_{cr}^m)\\
		&\ \ \; \; \; = \mathtt{DP}(\mathtt{FC}(\mathtt{DP}(\alpha(\mathtt{FC}(\mathrm{X}_{cr}^m))))),\\
		&\mathrm{X}_{fr}^m = \mathtt{LN}(\mathrm{X}^m + \mathrm{X}_{cr}^m + \mathrm{X}_{f}^m).
    \end{split}
\end{equation}

\subsection{Emotion Classifier}
After multiple layers of RUME and ACME encoding, we obtain the final feature matrices $H^T$, $H^V$, and $H^A$. Then they are concatenated to obtain fused feature matrix $H$. Finally, the feature dimensions of $H$ are converted to $|E|$ (number of emotions) with an emotion classifier, and thus we obtain the predicted emotion $e_i^\prime$ ($e_i^\prime \in E$). The process can be formulated as follows:
\begin{equation}
    \begin{split}
        &l_i = \mathtt{DP}(\mathtt{ReLU}(W_l h_i)),\\
		&y_i^\prime = \mathtt{SMAX}(W_{smax} l_i),\\
		&e_i^\prime = \mathtt{ARGMAX}(y_i^\prime[k]),
    \end{split}
\end{equation}
where $h_i \in H$; $W_l$ and $W_{smax}$ are learnable parameters; $\mathtt{ARGMAX}(\cdot)$ denotes the argmax function. We define the loss function as follows:
\begin{equation}
    \label{eq:loss_c}
    \mathcal{L}_c = - \frac {1}{\sum_{I=0}^{N-1} n(I)} \sum_{i=0}^{N-1}\sum_{j=0}^{n(i)-1} y_{ij} \log y_{ij}^\prime,
\end{equation}
where $n(i)$ is the number of utterances of the $i$-th dialogue, and $N$ is the number of all dialogues in training set; $y_{ij}^{\prime}$ denotes the probability distribution of predicted emotion label of the $j$-th utterance in the $i$-th dialogue, and $y_{ij}$ denotes the ground truth label.

\subsection{Label Based Emotion-Shift Module}
In order to extract emotion-shift information and enhance the emotional expression of the utterance, we introduce the label based emotion-shift module (LESM) to explicitly model the emotion-shift between utterances. LESM consists of three main steps, i.e., firstly, constructing the probability tensor of emotion-shift, then generating the label matrix of emotion-shift, and finally, performing the training exploiting the loss of emotion-shift. Our LESM is used as an auxiliary task to guide the learning of the main task, thereby empowering the main task to reduce intra-modal conceptual modeling during emotion shift scene and instead focus on cross-modal interactive modeling.
\begin{figure*}[htbp]
    \centering
    \includegraphics[width=7.1in]{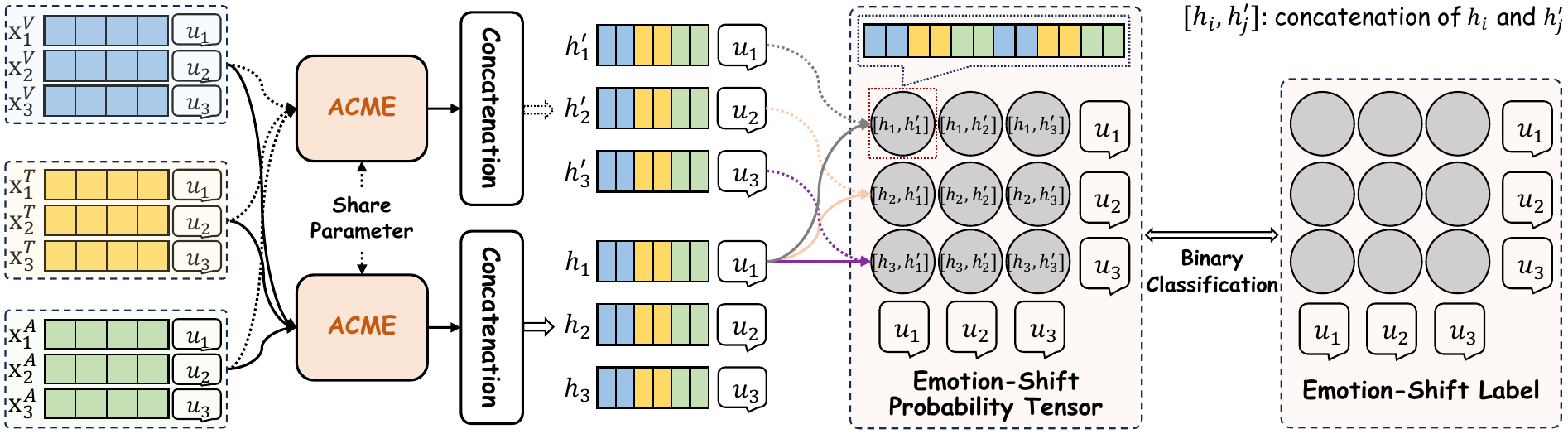}
    \caption{An example of constructing emotion-shift probability tensor $\mathcal{T}_{123}$. Here, $\mathcal{T}_{123}$ can be viewed as a $3 \times 3$ dimensional matrix composed of feature vectors (emotion-shift probability vectors) that are concatenated from the feature vectors of utterances.}
    \label{fig:emotion_shift}
\end{figure*}

\subsubsection{Emotion-Shift Probability}
Inspired by SimCSE~\cite{gao2021simcse}, we employ two parameter-shared ACMEs to generate two feature matrices with different representations but consistent emotion semantics. In other words, the output $\mathrm{X}^m$ ($m \in \{T,V,A\}$) of RUME is treated as the inputs to two parameter-shared ACMEs, and then two fused feature matrices $H$ and $H^\prime$ are obtained. Here, $H \in \mathbb{R}^{|U| \times |F|}$, $H^\prime \in \mathbb{R}^{|U| \times |F|}$, $|U|$ is the number of utterances in the conversation, and $|F|$ is feature dimension of $H$ or $H^\prime$. We concatenate the feature vectors from each utterance in $H$ and all utterances in $H^\prime$ to construct $|U|\times|U| \times 2|F|$ dimensional emotion-shift probability tensor $\mathcal{T}$. If the feature dimension of $\mathcal{T}$ is mapped to 1 through the fully-connected layer, then the emotion-shift probability between two utterances can be obtained.

An example of the above process can be illustrated in Fig.~\ref{fig:emotion_shift}. Specifically, assume that there exist three utterances and the corresponding feature vectors are $\mathrm{x}_1^m$, $\mathrm{x}_2^m$, and $\mathrm{x}_3^m$. These feature vectors are taken as inputs to two parameter-shared ACMEs, and thus the fused feature vectors ${h}_i$ and ${h}_i^\prime$ ($i=1,2,3$) are obtained, where ${h}_i \in H$ and ${h}_i^\prime \in H^\prime$. Then, concatenating ${h}_1$ with each ${h}_i^\prime$ (i.e., ${h}_1^\prime$, ${h}_2^\prime$, and ${h}_3^\prime$); and similarly, concatenating ${h}_2$ with each ${h}_i^\prime$; and for ${h}_3$, the same concatenation operation is adopted. Finally, the $3 \times 3 \times 2|F|$ dimensional emotion-shift probability tensor $\mathcal{T}_{123}$ is obtained.

\subsubsection{Emotion-Shift Label}
We annotate emotion-shift status between utterances based on true emotion labels of the dataset. Concretely, if the true emotions of two utterances are the same, then we annotate their shift status as 0, meaning that emotion shift has not occurred; conversely, if their true emotions are different, then we annotate the shift status as 1, meaning that emotion shift has occurred. By the above operation, we obtain the $|U|\times|U|$ dimensional emotion-shift label matrix.

\subsubsection{Emotion-Shift Loss} 
After constructing the emotion-shift probabilities and labels, we require to define the corresponding emotion-shift loss for training. LESM is a binary-classified auxiliary task, which aims to correctly distinguish the emotion-shift states between utterances. In this way, the model is prompted to capture emotion-shift information, thereby guiding it to attenuate focus on contextual information. First, in order to obtain the predicted emotion-shift state $s_{ij}^\prime$ ($s_{ij}^\prime \in \{0,1\}$), we convert the feature dimension of the probability tensor $\mathcal{T}$ to 2 with the fully-connected layer. The above process is as follows:
\begin{equation}
    \begin{split}
        &l_{ij}^\prime = \mathtt{DP}(\mathtt{ReLU}(W_l^\prime t_{ij})),\\
		&z_{ij}^\prime = \mathtt{SMAX}(W_{smaxs} l_{ij}^\prime),\\
		&s_{ij}^\prime = \mathtt{ARGMAX}(z_{ij}^\prime[k]),
    \end{split}
\end{equation}
where $t_{ij}$ denotes emotion-shift probability vector between the $i$-th and $j$-th utterances, $t_{ij} \in \mathcal{T}$; $z_{ij}^\prime$ is the probability distribution of predicted emotion-shift label between the $i$-th and $j$-th utterances; $W_l^\prime$ and $W_{smaxs}$ are learnable parameters. The defined emotion-shift loss is:
\begin{equation}
    \label{eq:loss_s}
	\begin{split}
    \mathcal{L}_s = - \frac {1}{\sum_{I=0}^{N-1} (n(I))^2} \sum_{i=0}^{N-1}\sum_{j=0}^{n(i)-1}
	\sum_{k=0}^{n(i)-1} z_{ijk} \log z_{ijk}^\prime,
	\end{split}
\end{equation}
where $n(i)$ is the number of utterances of the $i$-th dialogue, and $N$ is the number of all dialogues in training set; $z^{\prime}_{ijk}$ denotes the probability distribution of predicted emotion-shift label between the $j$-th and $k$-th utterances in the $i$-th dialogue, and $z_{ijk}$ denotes the ground truth label.

\subsection{Training Objective}
We combine the classification loss $\mathcal{L}_c$ and emotion-shift loss $\mathcal{L}_s$ to get the final training objective,
\begin{equation}
    \label{eq:loss}
    \mathcal{L} = \mathcal{L}_c + \lambda \mathcal{L}_s + \eta  \Vert W \Vert,
\end{equation}
where $\lambda$ is a trade-off parameter with a value in the range [0,1], $\eta$ is the L2-regularizer weight, and $W$ is the set of all learnable parameters. Further, $\lambda$ can be set manually or automatically adjusted using the method of Kendall et al.~\cite{kendall2018multi}.

\section{Experimental Setup}\label{setup} 
\subsection{Datasets}
We adopt two public dialogue emotion datasets: MELD~\cite{poria2018meld} and IEMOCAP~\cite{busso2008iemocap} . The statistics of them are shown in TABLE~\ref{tab:statistics}. 
\textbf{MELD} is a multimodal and multiparty dataset containing more than 1,400 dialogues and 13,000 utterances from the TV series \textit{Friends}. There are seven emotion labels in the dataset, i.e., \textit{anger}, \textit{disgust}, \textit{sadness}, \textit{joy}, \textit{surprise}, \textit{fear}, and \textit{neutral}. 1,153 dialogues with 11,098 utterances are employed as the training and validation sets, where the 10\% of utterances is selected as the validation set. The remaining 2,610 utterances in the dataset are served as the test set, which contains 280 dialogues. 
\textbf{IEMOCAP} is an acted, multimodal and multi-speaker dataset consisting of dyadic conversations, which contains textual, visual, and acoustic modalities. The dataset consists of 151 dialogues and 7,433 utterances labelled with six emotion categories: \textit{happy}, \textit{sad}, \textit{neutral}, \textit{angry}, \textit{excited}, and \textit{frustrated}. We adopt 120 dialogues with 5,810 utterances for training and validation, and the rest for testing. Here, the validation set is randomly selected from the training set with a ratio of 10\%.
\begin{table}[htbp]
    \centering
    \renewcommand{\arraystretch}{1.0}
    \setlength{\tabcolsep}{4pt}
    \caption{The Statistical Information of MELD and IEMOCAP}
    \begin{threeparttable}
    \begin{tabular}{c|cccc}
    \hline
    \multirow{2}{*}{Datasets} &\multicolumn{2}{c}{MELD} &\multicolumn{2}{c}{IEMOCAP} \\
           &\#Dialogue &\#Utterance &\#Dialogue &\#Utterance \\ 
    \hline
    Train &1,039 &9,989 &100 &5,163  \\
    Val &114 &1,109 &20 &647 \\
    Test &280 &2,610 &31 &1,623 \\
    \hline
    \end{tabular}
    \#Dialogue and \#Utterance denote the number of dialogues and utterances, respectively.
    \end{threeparttable}
    \label{tab:statistics}
\end{table}

The utterance-level features are extracted in the following manner. The visual and acoustic features are extracted with the way of MMGCN~\cite{hu2021mmgcn}, i.e., the visual features are extracted using a DenseNet~\cite{huang2017densely} pre-traind on the Facial Expression Recognition Plus corpus\cite{barsoum2016training}, the acoustic features are extracted using the OpenSmile toolkit with IS10 configuration~\cite{schuller2011recognising}. The textual feature is processed adopting the approach of COSMIC~\cite{ghosal2020cosmic}, i.e., the RoBERTa~\cite{liu2019roberta} model is applied for pre-training and fine-tuning to extract textual features.

\subsection{Training Details}
The operating system we used is Ubuntu with version 20.04, and the deep learning framework is Pytorch 2.0.0. All experiments are conducted on a single NVIDIA GeForce RTX 3090. In our experiments, the maximum epoch is set to 80, and the basis network of RUME is GRU by default; AdamW~\cite{loshchilov2018decoupled} is employed as the optimizer with the L2 regularization factor of 1e-4; and the number of heads in all MHA networks is set to 8. For the MELD dataset, the learning rate is set to 1e-5, and the batch size is set to 64; the number of network layers for RUME and ACME are 2 and 3, respectively, with corresponding dropout rates of 0.1 and 0.3, respectively; we manually set the trade-off parameter $\lambda$ to 0.9 by default. For the IEMOCAP dataset, the learning rate is set to 2e-5, and the batch size is set to 32; the number of network layers for RUME and ACME are 2 and 5, respectively, with corresponding dropout rates of 0.2 and 0.4, respectively; the trade-off parameter $\lambda$ is manually set to 1.0. 

\subsection{Comparative Methods and Evaluation Metrics}
The baselines we use are divided into two categories: text based methods and multi-modality based methods. The text based approaches include AGHMN~\cite{jiao2020real}, DialogXL~\cite{shen2021dialogxl}, I-GCN~\cite{nie2022igcn}, CauAIN~\cite{zhao2022cauain}, CoG-BART~\cite{li2022contrast}. The multi-modality based approaches include MMGCN~\cite{hu2021mmgcn}, DialogueTRM~\cite{mao2021dialoguetrm}, MetaDrop~\cite{chen2021learning}, HU-Dialogue~\cite{chen2022modeling}, MM-DFN~\cite{hu2022mmdfn}, COGMEN~\cite{joshi2022cogmen}, UniMSE~\cite{hu2022unimse}.

Following previous works~\cite{hu2021mmgcn,hu2022mmdfn}, we report the accuracy (Acc) and weighted F1 score (W-F1) to measure overall performance on these two public datasets (i.e., MELD and IEMOCAP), and also present F1 score for each emotion class.

\section{Results and Analysis}\label{result}
\subsection{Comparison to Baselines on the MELD Dataset}
\begin{table*}[htbp]
    \centering
    \renewcommand{\arraystretch}{1.0}
    \footnotesize
    \setlength{\tabcolsep}{3pt}
    \caption{Overall Results of All Models on the MELD Dataset}
	\begin{threeparttable}
    \begin{tabular}{c|ccccccc|cc}
    \hline
    \multirow{3}{*}{Models} & \multicolumn{9}{c}{MELD} 
	\\ \cline{2-10} 
    & \textit{neutral} & \textit{surprise} &\textit{fear} & \textit{sadness} & \textit{joy} &\textit{disgust} &\textit{anger}  & \multirow{2}{*}{W-F1} & \multirow{2}{*}{Acc} \\ 
    & F1 & F1  & F1 & F1 & F1 & F1 & F1 &  &  \\
    \hline
	AGHMN~\cite{jiao2020real} & 76.40  & 49.70 & 11.50 & 27.00 & 52.40 & 14.00 & 39.40 & 58.10 &63.50\\
    DialogXL~\cite{shen2021dialogxl}  & - & - & - & - & - & - & - &62.41 &- \\
    I-GCN~\cite{nie2022igcn} &   78.00 & 51.60 & 8.00 & 38.50 & 54.70 & 11.80 & 43.50 &60.80 &- \\
    CauAIN~\cite{zhao2022cauain}    & - & - & - & - & - & - & - &65.46 &- \\
    CoG-BART~\cite{li2022contrast}   & -  & - & - & - & - & - & - &64.81 &- \\
	\hline
    MMGCN~\cite{hu2021mmgcn}  & 76.33 & 48.15 & - & 26.74 & 53.02 & - & 46.09 &58.31 &60.42 \\
    DialogueTRM~\cite{mao2021dialoguetrm}  & - & - & - & - & - & - & - &63.50 &65.70\\
    MetaDrop~\cite{chen2021learning}  & - & - & - & - & - & - & -  &66.08 &66.42 \\
	HU-Dialogue~\cite{chen2022modeling}  & - & - & - & - & - & - & -  &58.56 &61.38 \\
	MM-DFN~\cite{hu2022mmdfn}  & 77.76 & 50.69 & - & 22.93 & 54.78 & - & 47.82  &59.46 &62.49 \\
	UniMSE~\cite{hu2022unimse}  & - & - & - & - & - & - & -  &65.51 &65.09 \\
	\hline
	\multirow{2}{*}{CFN-ESA}  & \textbf{80.05}$^\dagger$ & \textbf{58.78}$^\dagger$ & \textbf{21.62}$^\dagger$  & \textbf{41.82}$^\dagger$ & \textbf{66.50}$^\dagger$ & \textbf{26.92}$^\dagger$ & \textbf{54.18}$^\dagger$ & \textbf{66.70}$^\dagger$ &\textbf{67.85}$^\dagger$ \\
     &79.93$\pm$0.40$^\ddagger$ &58.47$\pm$0.37$^\ddagger$ &22.41$\pm$2.24$^\ddagger$ &41.16$\pm$2.23$^\ddagger$ &64.78$\pm$1.25$^\ddagger$ &30.14$\pm$2.50$^\ddagger$ &53.91$\pm$1.25$^\ddagger$ &66.36$\pm$0.27$^\ddagger$ &67.42$\pm$0.32$^\ddagger$  \\
      \hline
    \end{tabular}%
    Results for MMGCN are from MM-DFN, other results are from the original papers. W-F1, F1, and Acc denote the accuracy (\%), F1 score (\%), and weighted F1 score (\%), respectively. The marker $^\dagger$ indicates the best result from the five experiments, and the marker $^\ddagger$ denotes the confidence interval.
    \end{threeparttable}
    \label{tab:meld_results}%
\end{table*}%
We report the experimental results of CFN-ESA on the MELD dataset in TABLE~\ref{tab:meld_results}. As can be seen from the table, the proposed CFN-ESA outperforms the results of all the baseline models in terms of weighted F1 score and accuracy. Among all the textual unimodal models, the weighted F1 score of CauAIN is 65.46\%, which is the highest experimental performance. Our CFN-ESA is 66.70\%, which is an improvement of 1.24\% relative to CauAIN. This result suggests that the acoustic and visual modalities in CFN-ESA can contribute complementary information to effectively improve the performance of the model. Relative to MetaDrop's weighted F1 score of 66.08\%, the proposed CFN-ESA improves by 0.62\%. The accuracy of MetaDrop is 66.42\%, while that of CFN-ESA is 67.85\%, with the former being 1.43\% lower than the latter. Comparing the accuracy of CFN-ESA and DialogueTRM, the accuracy of CFN-ESA improves by 2.15\% relative to that of DialogueTRM, yielding similar results as above. These comparative results indicate that our model can more effectively model multimodal emotion datasets.

\begin{figure}[htbp]
    \centering
    \includegraphics[width=3.4in]{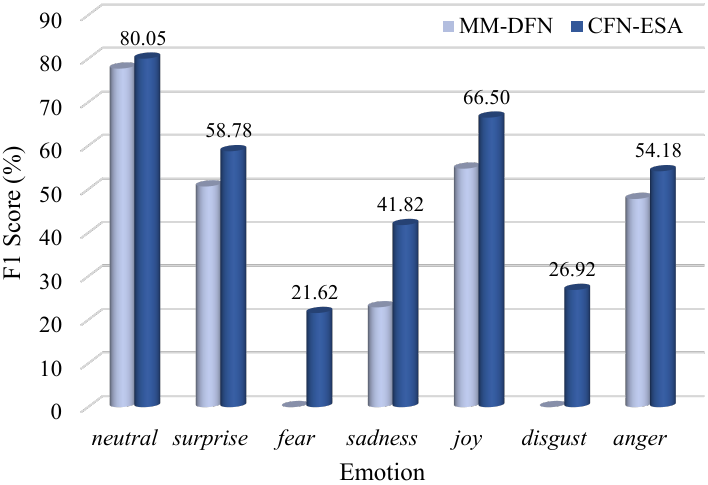}
    \caption{F1 scores of CFN-ESA and MM-DFN for each emotion class.}
    \label{fig:meld_per_f1}
\end{figure}
As can be noticed from TABLE~\ref{tab:meld_results}, our CFN-ESA achieves F1 scores of 21.62\% and 26.92\% on these two emotions, which are significantly higher than the results of AGHMN and I-GCN. We show the F1 scores of CFN-ESA and MM-DFN for each emotion class in Fig.~\ref{fig:meld_per_f1}. It is obvious that our CFN-ESA outperforms MM-DFN in the experimental results for all emotion classes. CFN-ESA achieves the best F1 score of 80.05\% for \textit{neutral} relative to all other emotions. Of particular concern is that the MELD dataset has an extremely severe class imbalance problem, where the emotions \textit{fear} and \textit{disgust} belong to the minority classes among all the classes. 

\subsection{Comparison to Baselines on the IEMOCAP Dataset}
\begin{figure}[htbp]
    \centering
    \subfloat[Before Feature Extraction]{\includegraphics[height=1.7in]{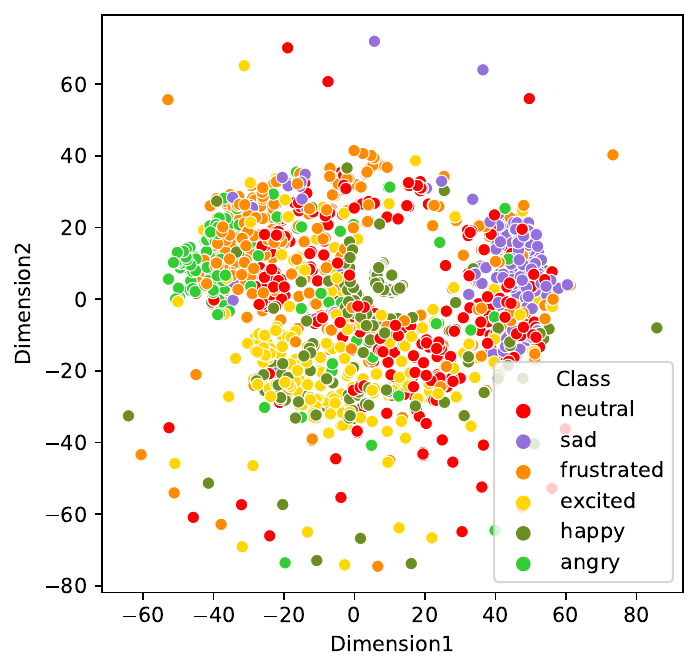}%
    \label{fig:iemocap_initial_emo}}
    \hfil
    \subfloat[After Feature Extraction]{\includegraphics[height=1.7in]{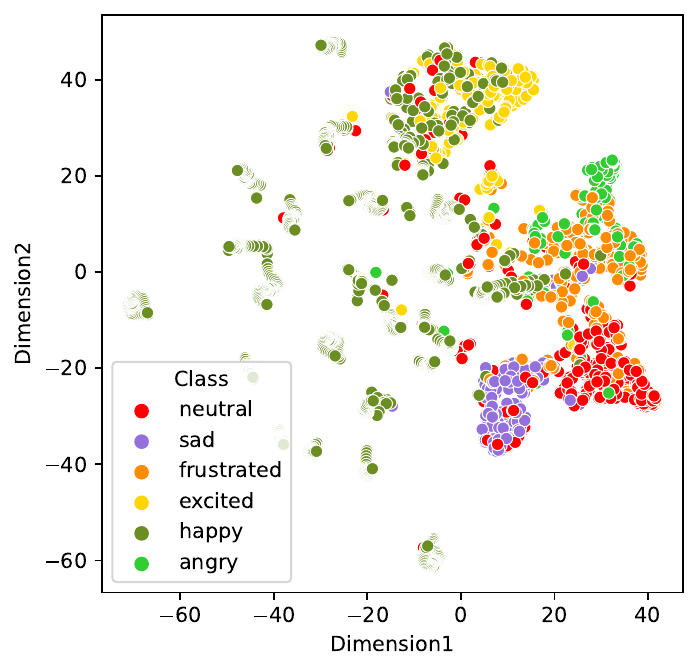}%
    \label{fig:iemocap_cfnesa_emo}}
    \caption{Comparison of T-SNE visualization before and after feature extraction is performed by employing CFN-ESA on the IEMOCAP dataset.}
    \label{fig:iemocap_tsne_emo}
\end{figure}

The comparison results of CFN-ESA on the IEMOCAP dataset are reported in TABLE~\ref{tab:iemocap_results}. We can state that CFN-ESA achieves the best performance with the weighted F1 score and accuracy of 71.04\% and 70.78\%, respectively. Focusing our attention on the comparison with the unimodal approaches. Relative to the weighted F1 score of 66.18\% for CoG-BART, that for CFN-ESA has an improvement of 4.86\%. The accuracy of I-GCN is 65.50\%, which is 5.28\% lower than that of our CFN-ESA. This phenomenon can indicate that CFN-ESA effectively leverages the information from multiple modalities and alleviates the problem of insufficient information expression in unimodal models. In the multimodal methods, our CFN-ESA still shows a strong performance. The weighted F1 score of UniMSE is 70.66\%, which is 0.38\% lower than the result of the proposed CFN-ESA. The performance of CFN-ESA improves by 1.77\% relative to the 69.01\% accuracy of MetaDrop. From these, we can conclude that CFN-ESA can more adequately capture multimodal complementary information in comparison to previous multimodal methods.
\begin{table*}[htbp]
    \centering
    \renewcommand{\arraystretch}{1.0}
    \footnotesize
    \setlength{\tabcolsep}{5pt}
    \caption{Overall Results of All Models on the IEMOCAP Dataset}
    \begin{tabular}{c|cccccc|cc}
    \hline
    \multirow{3}{*}{Models} & \multicolumn{8}{c}{IEMOCAP} 
	\\ \cline{2-9} 
    &  \textit{happy} & \textit{sad} & \textit{neutral} & \textit{angry} &\textit{excited} &\textit{frustrated}  & \multirow{2}{*}{W-F1} & \multirow{2}{*}{Acc} \\ 
    &  F1 & F1  & F1 & F1 & F1 & F1 &  &  \\
    \hline
	AGHMN~\cite{jiao2020real} & 52.10  & 73.30 & 58.40 & 61.90 & 69.70 & 62.30 & 63.50 &63.50\\
    DialogXL~\cite{shen2021dialogxl}  & - & - & - & - & - & - &62.41 &- \\
    I-GCN~\cite{nie2022igcn}  & 50.00 & \textbf{83.80} & 59.30 & 64.60 & 74.30 & 59.00 &65.40 &65.50 \\
    CauAIN~\cite{zhao2022cauain}  & - & - & - & - & - & - &67.61 &- \\
    CoG-BART~\cite{li2022contrast}  & -  & - & - & - & - & - &66.18 &- \\
	\hline
    MMGCN~\cite{hu2021mmgcn}  & 45.14 & 77.16 & 64.36 & 68.82 & 74.71 & 61.40 &66.26 &66.36 \\
    DialogueTRM~\cite{mao2021dialoguetrm}  & - & - & - & - & - & - &69.70 &69.50\\
    MetaDrop~\cite{chen2021learning} & - & - & - & - & - & -  &69.04 &69.01 \\
	HU-Dialogue~\cite{chen2022modeling} & - & - & - & - & - & -  &65.36 &65.72 \\
	MM-DFN~\cite{hu2022mmdfn} & 42.22 & 78.98 & 66.42 & 69.77 & \textbf{75.56} & 66.33  &68.18 &68.21 \\
	COGMEN~\cite{joshi2022cogmen} & 51.90 & 81.70 & 68.60 & 66.00 & 75.30 & 58.20 & 67.60 &68.20 \\
	UniMSE~\cite{hu2022unimse} & - & - & - & - & - & -  &70.66 &70.56 \\
	\hline
	\multirow{2}{*}{CFN-ESA}  & \textbf{53.67}$^\dagger$ & 80.60$^\dagger$  & \textbf{71.65}$^\dagger$ & \textbf{70.32}$^\dagger$ & 74.82$^\dagger$ & \textbf{68.06}$^\dagger$ & \textbf{71.04}$^\dagger$ &\textbf{70.78}$^\dagger$ \\
    &56.76$\pm$2.60$^\ddagger$ &81.34$\pm$0.62$^\ddagger$ &71.19$\pm$0.63$^\ddagger$ &68.23$\pm$2.98$^\ddagger$ &75.83$\pm$1.71$^\ddagger$ &65.50$\pm$3.19$^\ddagger$ &70.69$\pm$0.29$^\ddagger$ &70.61$\pm$0.24$^\ddagger$ \\
      \hline
    \end{tabular}%
    \label{tab:iemocap_results}%
\end{table*}%
Compared to MM-DFN, CFN-ESA achieves superior performance on all emotions except \textit{excited}. Particularly, it is evident that the proposed CFN-ESA achieves an F1 score of 53.67\% for the emotion \textit{happy}, which is significantly higher than the 42.22\% of MM-DFN. The F1 score of CFN-ESA in terms of \textit{neutral} is improved by 5.23\% than that of MM-DFN. In addition, we can observe from TABLE~\ref{tab:iemocap_results} that \textit{sad} achieves the highest F1 scores among all the emotion classes. Fig.~\ref{fig:iemocap_tsne_emo} shows the T-SNE visualization of the original feature and the feature extracted by CFN-ESA on the IEMOCAP dataset. It can be observed that the feature extracted by CFN-ESA can clearly distinguish each emotion class and outperform the original feature, demonstrating the powerful capability of our model for feature extraction.

\subsection{Effect of Different Modal Settings}
As shown in TABLE~\ref{tab:modal_settings}, we examine the effects of different modal settings on the proposed model. Specifically, the input configurations for the three shared-parameter RUMEs are as follows: (1) when exploiting three modalities, the inputs to three RUMEs are $X^T$, $X^V$, and $X^A$, where $T$, $V$, and $A$ denote the textual, visual, and acoustic modalities, respectively; (2) when the textual and visual modalities are utilized, the inputs to three RUMEs are $X^T$, $X^V$, and $X^V$, respectively; (3) when engaging with both textual and acoustic modalities, the inputs to three RUMEs are $X^T$, $X^A$, and $X^A$, respectively; (4) In the case of utilizing the visual and acoustic modalities exclusively, the inputs to three RUMEs are $X^V$, $X^A$, and $X^A$, respectively; and (5) when working with a single modality, the inputs to all three RUMEs are $X^m$, where $m \in \{T, V, A\}$.
\begin{table}[htbp]
    \centering
    \renewcommand{\arraystretch}{1.0}
    \setlength{\tabcolsep}{8pt}
    \caption{Performance comparison of Different Modal Settings}
    \begin{threeparttable}
    \begin{tabular}{c|cc|cc}
    \hline
    \multirow{2}{*}{Modal Settings} &\multicolumn{2}{c|}{MELD} &\multicolumn{2}{c}{IEMOCAP}\\
           &W-F1 &Acc &W-F1 &Acc\\ 
    \hline 
	Textual &65.81 &67.09 &66.57 &66.56 \\
    Visual &32.05 &48.05 &44.23 &45.01 \\
	Acoustic &41.46 &49.35 &51.08 &53.45 \\
	\hline
	T + V &65.93 &67.13 &67.86 &67.58 \\
    T + A &65.94 &67.16 &68.46 &68.67 \\
	V + A &43.25 &50.34 &59.83 &60.36 \\
	\hline
    T + V + A &\textbf{66.70} &\textbf{67.85} &\textbf{71.04} &\textbf{70.78} \\
	\hline
    \end{tabular}
    T, V, and A is textual, visual, and acoustic modalities, respectively.
    \end{threeparttable}
    \label{tab:modal_settings}
\end{table}

As we expected, the tri-modal setting achieves the best performance relative to the bi-modal and unimodal settings. Among all the unimodal settings, the textual setting attains 67.09\% accuracy on the MELD dataset and 66.57\% F1 score on the IEMOCAP dataset, which is much higher than two other unimodal settings and reaches the best performance. These results indicate that the textual modality contains more emotional information than other two modalities. Compared to visual unimodal setting, the acoustic unimodal setting yields better experimental results on both datasets. The plausible explanation for this observation is that the image often incorporate a more intricate background and are susceptible to a higher degree of ambient noise interference.

\begin{figure}[htbp]
    \centering
    \includegraphics[width=3.4in]{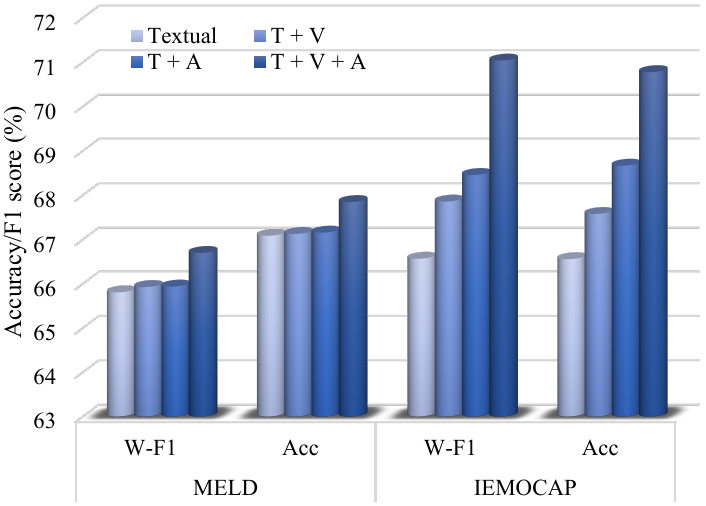}
    \caption{Comparison of different modal settings with textual modality.}
    \label{fig:different_modal_settings}
\end{figure}

The performance of the bi-modal settings with text is better compared to the visual-acoustic setting. On the IEMOCAP dataset, the textual-acoustic setting achieves an accuracy of 68.67\%, which is 1.09\% higher than the result of the textual-visual setting. Similar experimental results also appear on the MELD dataset. In addition, Fig.~\ref{fig:different_modal_settings} illustrates the comparison among the textual unimodal, textual-visual bi-modal, textual-acoustic bi-modal, and tri-modal settings. It can be observed that the bi-modal setting with visual or acoustic modality has a higher performance than the textual setting. This indicates that the multimodal settings can effectively improve the performance of the ERC task. Similarly, the experimental results of the tri-modal setting with both visual and acoustic modalities are better compared to the bi-modal setting.

\subsection{Impact of Different Network Depths}
\begin{figure*}[htbp]
    \centering
    \subfloat[Effect of RUME]{\includegraphics[height=2.0in]{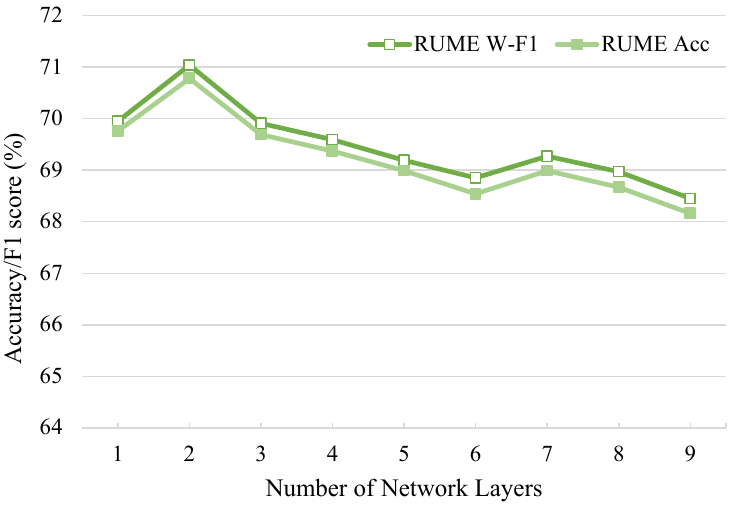}%
    \label{fig:rume_depths}}
    \hfil
    \subfloat[Effect of ACME]{\includegraphics[height=2.0in]{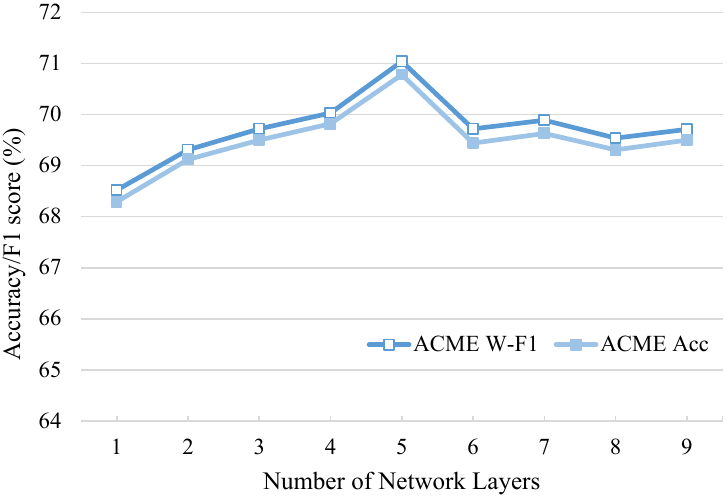}%
    \label{fig:acme_depths}}
    \caption{Effect of different network depths on the performance. The subfigure on the left (or right) indicates the effect of network depth for RUME (or ACME).}
    \label{fig:depths}
\end{figure*}

We explore the effect of different network depths (number of layers) on the performance in this subsection. We first fix the network depth of one encoder unchanged, then vary the network depth of the other, and record the experimental results. Note that these experiments are conducted on the IEMOCAP dataset. Fig.~\ref{fig:rume_depths} depicts the effect of RUME with different network depths on the experimental results. As evidenced by the depicted figure, it is discernible that the performance of CFN-ESA initially ascends and subsequently descends with an increase in network depth, reaching its peak at a depth of 2 layers. The impact of ACME with varying network depths on our model is illustrated in Fig.~\ref{fig:acme_depths}, which reveals a similar trend to that observed in Fig.~\ref{fig:rume_depths}. Specifically, the experimental outcomes exhibit a pattern of escalation followed by attenuation, with the optimal network depth being 5 layers.

\subsection{Impact of Different Trade-Off Parameters}
In our experiments, the trade-off parameter $\lambda$ can be set in two ways, that is, manual setting and automatic setting using the method of Kendall et al.~\cite{kendall2018multi}. In this subsection, we investigate the effects of different trade-off parameters on the performance. Table~\ref{tab:trade-off} demonstrates the effect of $\lambda$ on the results on the MELD and IEMOCAP datasets. It can be seen that: (1) on the MELD dataset, the best experimental results are achieved when $\lambda$ is manually set to 0.9; and (2) on the IEMOCAP dataset, the best weighted F1 score is attained when $\lambda$ is manually set to 1.0, whereas automatically setting $\lambda$ results in the best accuracy.
\begin{table}[htbp]
    \centering
    \renewcommand{\arraystretch}{1.0}
    \setlength{\tabcolsep}{9pt}
    \caption{Performance Comparison of Different Trade-Off Parameters}
    \begin{tabular}{cc|cc|cc}
    \hline
    \multicolumn{2}{c|}{\multirow{2}{*}{Values of ${\lambda}$}} &\multicolumn{2}{c|}{MELD} &\multicolumn{2}{c}{IEMOCAP}\\ 
           & &W-F1 &Acc &W-F1 &Acc\\ 
    \hline 
	\multicolumn{1}{c|}{\multirow{10}{*}{Manual}} & 0.1 &66.51 &67.74 &70.77 &70.52 \\
    \multicolumn{1}{c|}{}& 0.2 &66.47 &67.70 &70.64 &70.40 \\
	\multicolumn{1}{c|}{}& 0.3 &66.53 &67.74 &70.77 &70.52 \\
	\multicolumn{1}{c|}{}& 0.4 &66.52 &67.74 &70.78 &70.52 \\
	\multicolumn{1}{c|}{}& 0.5 &66.53 &67.78 &70.64 &70.40 \\
	\multicolumn{1}{c|}{}& 0.6 &66.58 &67.78 &70.65 &70.40 \\
	\multicolumn{1}{c|}{}& 0.7 &66.57 &67.74 &70.75 &70.52 \\
	\multicolumn{1}{c|}{}& 0.8 &66.67 &67.82 &70.90 &70.65 \\
    \multicolumn{1}{c|}{}& 0.9 &\textbf{66.70} &\textbf{67.85} &70.96 &70.72 \\
	\multicolumn{1}{c|}{}& 1.0 &66.68 &67.82 &\textbf{71.04} &70.78 \\
	\hline
	\multicolumn{2}{c|}{Automatic} &66.64 &67.78 &70.72 &\textbf{70.98}\\
	\hline
    \end{tabular}
    \label{tab:trade-off}
\end{table}

\subsection{Ablation Studies}
To demonstrate the effectiveness of each module in CFN-ESA, we perform a series of ablation experiments in this subsection. Specifically, we remove the recurrence based uni-modality encoder (RUME), attention based cross-modality encoder (ACME), and label based emotion-shift module (LESM), respectively, then report the experimental results. The results are showed in TABLE~\ref{tab:ablation}.
\begin{table}[htbp]
    \centering
    \renewcommand{\arraystretch}{1.0}
    \setlength{\tabcolsep}{10pt}
    \caption{Performance Comparison After Removing Each Module}
    \begin{threeparttable}
    \begin{tabular}{c|cc|cc}
    \hline
    \multirow{2}{*}{Models} &\multicolumn{2}{c|}{MELD} &\multicolumn{2}{c}{IEMOCAP}\\
           &W-F1 &Acc &W-F1 &Acc\\ 
    \hline 
	CFN-ESA &\textbf{66.70} &\textbf{67.85} &\textbf{71.04} &\textbf{70.78} \\
	\hline
	-w/o RUME &66.37 &67.51 &70.25 &70.33 \\
    -w/o ACME &65.97 &67.20 &68.08 &67.97 \\
	-w/o LESM &66.40 &67.62 &70.22 &70.01 \\
	\hline
    \end{tabular}
    The markers -w/o RUME, -w/o ACME, and -w/o LESM denote removing RUME, ACME, and LESM, respectively.
    \end{threeparttable}
    \label{tab:ablation}
\end{table}
\begin{figure}[htbp]
    \centering
    \includegraphics[width=3.4in]{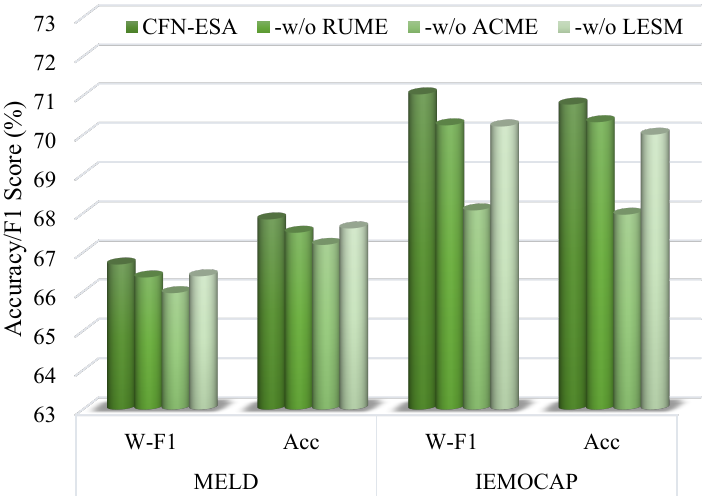}
    \caption{Visualization results for removing different modules on the performance.}
    \label{fig:ablation}
\end{figure}

Validity of RUME: When our RUME is removed, the weighted F1 score of the proposed model on the MELD dataset decreases from 66.70\% to 66.37\%; while on the IEMOCAP dataset, the accuracy of CFN-ESA decreases by 0.45\% from the original 70.78\%. The primary reason for the declines is that CFN-ESA loses the ability to model local context. Thus, our CFN-ESA relies on RUME to extract dialogue-level contextual information.

Validity of ACME: Since the input to LESM depends on two forward propagations of ACME, the input comes from the results of two forward propagations of RUME when ACME is removed. To put it differently, we directly use RUME instead of ACME. As can be seen from the table, when we remove ACME, the accuracy of our CFN-ESA on the MELD dataset decreases by 0.65\%, obtaining a result of 67.20\%; while on the IEMOCAP dataset, the model's weighted F1 scores show a significant decrease of 2.96\%. The above results indicate that our ACME plays an essential role in adequately capturing multimodal complementary information and has the capability to cross-modal interaction.

Validity of LESM: In similar fashion to the experimental results discussed previously, both the weighted F1 score and accuracy of the proposed CFN-ESA decline when we remove LESM. On the MELD dataset, the weight F1 score of our model drops to 66.40\%; on the IEMOCAP dataset, the accuracy of CFN-ESA decreases by 0.77\% from 70.78\%. These phenomena suggest that LESM, as an auxiliary task to ERC, can capture emotion-shift information in conversations, which facilitates the optimization and enhancement of emotional expression for utterances.

Overall, regardless of which module of CFN-ESA is removed, there are degradation in the performance on these two datasets. It can be visualized in Fig.~\ref{fig:ablation} that the performance of CFN-ESA decreases after the removal of different modules. In summary, it can be stated that these modules we designed for the model are valid.

\subsection{Comparison of ACME and Transformer Encoder}
\begin{figure}[htbp]
    \centering
    \subfloat[TFE-1]{\includegraphics[height=1.4in]{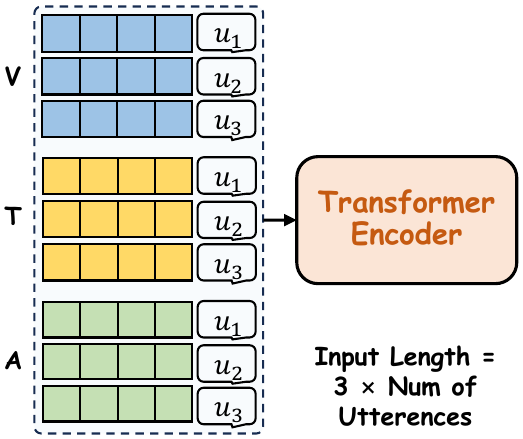}%
    \label{fig:tfe_1}}
    \hfil
    \subfloat[TFE-2]{\includegraphics[height=1.4in]{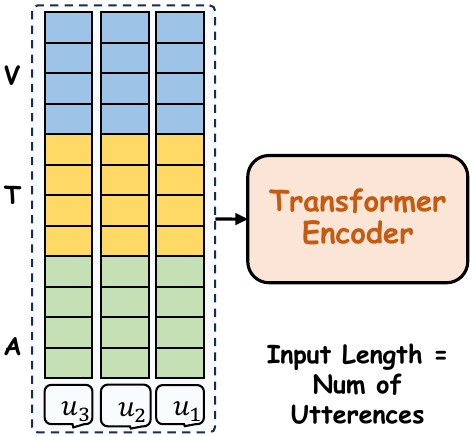}%
    \label{fig:tfe_2}}
    \caption{Illustration of two input schemes (i.e., TFE-1 and TFE-2) for TFE.}
    \label{fig:tfe}
\end{figure}

In this subsection, we attempt to replace ACME with a Transformer encoder (TFE)~\cite{vaswani2017attention}, employing two distinct input schemes: TFE-1 and TFE-2. They are defined as follows: (1) in TFE-1, the three uni-modal representations derived from RUME are combined at the sequence level, resulting in an input sequence length of $3 \times NumUtter$ and a feature dimension of $DimFeat$; (2) in TFE-2, the three uni-modal representations from RUME are concatenated at the feature level, yielding an input sequence length of $NumUtter$ and a feature dimension of $3 \times DimFeat$. Here, $NumUtter$ denotes the number of utterances, and $DimFeat$ represents the feature dimension for each utterance. As illustrated in Fig.~\ref{fig:tfe}, if there are 3 utterances with a feature dimension of 4 per utterance, then under these schemes, (1) TFE-1 has a sequence length of 9 and a feature dimension of 4; (2) TFE-2 has a sequence length of 3 and a feature dimension of 12. 

TABLE~\ref{tab:acme_tfe} reports the experimental results of these two schemes on the MELD and IEMOCAP datasets. On the MELD dataset, the scheme TFE-2 obtains an F1 score of 66.25\%, which is superior to TFE-1. The opposite result appears on the IEMOCAP dataset, with scheme TFE-1 achieving a higher F1 score compared to TFE-2. Regardless, ACME consistently surpasses both TFE schemes in terms of performance, indicating that our proposed ACME exhibits superior multimodal modeling capabilities over TFE.
\begin{table}[htbp]
    \centering
    \renewcommand{\arraystretch}{1.0}
    \setlength{\tabcolsep}{10pt}
    \caption{Performance Comparison of ACME and Transformer Encoder}
    \begin{threeparttable}
    \begin{tabular}{c|cc|cc}
    \hline
    \multirow{2}{*}{Modules} &\multicolumn{2}{c|}{MELD} &\multicolumn{2}{c}{IEMOCAP}\\
           &W-F1 &Acc &W-F1 &Acc\\ 
	\hline
	TFE-1 &65.91 &67.01 &68.74 &68.67 \\
    TFE-2 &66.25 &67.55 &68.25 &69.25 \\
    \hline 
	ACME &\textbf{66.70} &\textbf{67.85} &\textbf{71.04} &\textbf{70.78} \\
	\hline
    \end{tabular}
    \end{threeparttable}
    \label{tab:acme_tfe}
\end{table}

\subsection{Correlation Analysis}
\begin{figure*}[htbp]
    \centering
	\subfloat[F1 Scores on MELD]{\includegraphics[height=2.0in]{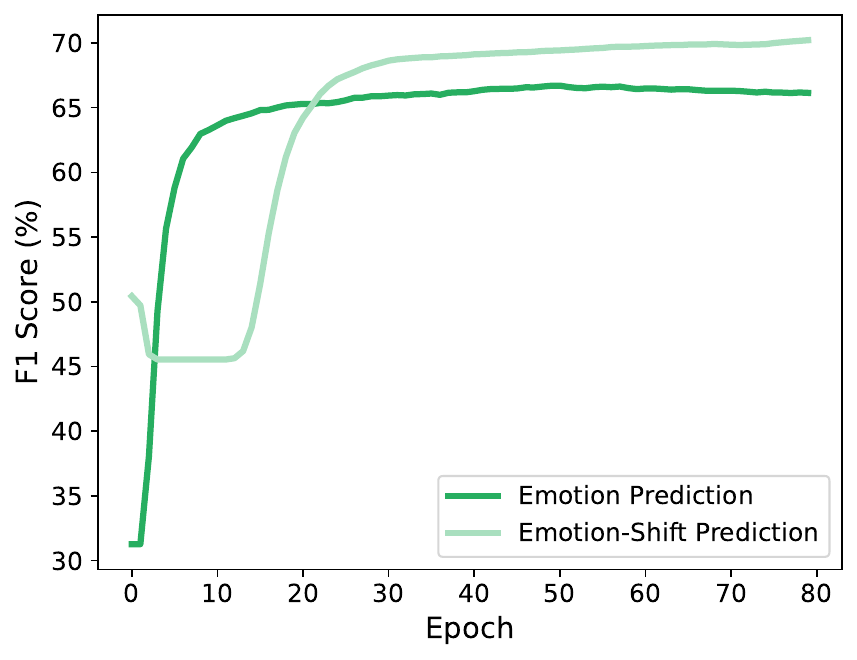}%
    \label{fig:meld_f1_epoch}}
    \hfil
    \subfloat[F1 Scores on IEMOCAP]{\includegraphics[height=2.0in]{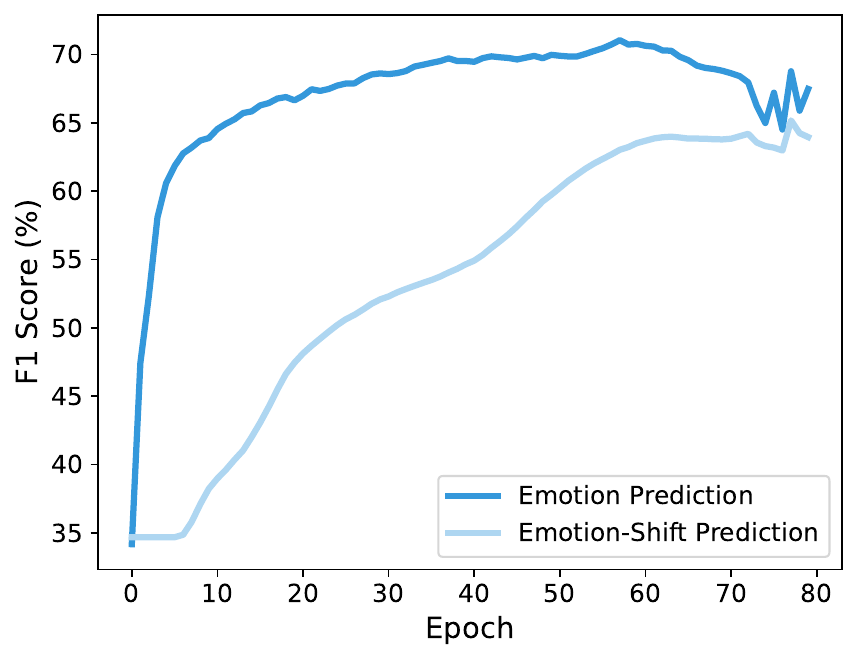}%
    \label{fig:iemocap_f1_epoch}}
    \caption{The variations in F1 scores for both emotion-shift prediction and emotion prediction as the epoch increases on the MELD and IEMOCAP datasets.}
    \label{fig:f1_epoch}
\end{figure*}

In Fig.~\ref{fig:f1_epoch}, we display the variations in F1 scores for both emotion-shift prediction and emotion prediction, with the aim to explore their inherent correlation. On the MELD dataset, as shown in Fig.~\ref{fig:meld_f1_epoch}, the scores for both emotion-shift prediction and emotion prediction are gradually increasing simultaneously as the epoch grows. As can be noticed from Fig.~\ref{fig:iemocap_f1_epoch}, a parallel trend is also evident on the IEMOCAP dataset. According to the above phenomena, it can be inferred that there exists a direct correlation between emotion-shift prediction and emotion prediction tasks. This means that emotion-shift prediction (auxiliary task) can promote emotion prediction (main task), which further validates the importance of LESM. In addition, these observations can inspire future methods to focus on improving the performance of the emotion-shift module, making the emotion prediction more accurate.

\subsection{Sentiment Classification}
We replace emotion with sentiment in this subsection in order to conduct the task of sentiment classification in conversations. In other words, we transform CFN-ESA into a three-classification (i.e., \textit{neutral}, \textit{positive}, and \textit{negative}) model. Note that since the IEMOCAP dataset does not contain sentiment labels, we need to merge the original emotion. The specific merging scheme is as follows: \textit{sad}, \textit{angry}, and \textit{frustrated} are merged into \textit{negative}; \textit{happy} and \textit{excited} are merged into \textit{positive}; \textit{neutral} remains unchanged.

The experimental results of our sentiment classification are reported in TABLE~\ref{tab:sentiment}. It can be observed that after the emotions are coarsened into sentiments, the weighted F1 scores and accuracies of CFN-ESA on these two datasets are improved. For instance, the accuracy of CFN-ESA on the MELD dataset is improved from 67.85\% to 73.75\%, with an increase of 5.9\%; on the IEMOCAP dataset, the weighted F1 score of the proposed CFN-ESA improves from 71.04\% to 84.49\%, with an increment of 13.45\%. 
\begin{table*}[htbp]
    \centering
    \renewcommand{\arraystretch}{1.0}
    \setlength{\tabcolsep}{7pt}
    \caption{Experimental Results for Sentiment Classification on these Two Datasets}
    \begin{threeparttable}
    \begin{tabular}{c|ccc|cc||ccc|cc}
    \hline
    \multirow{2}{*}{Models} &\multicolumn{5}{c||}{MELD} &\multicolumn{5}{c}{IEMOCAP}\\
    \cline{2-11}
           &\textit{neutral} &\textit{positive} &\textit{negative} &\multirow{2}{*}{W-F1} &\multirow{2}{*}{Acc} &\textit{neutral} &\textit{positive} &\textit{negative} &\multirow{2}{*}{W-F1} &\multirow{2}{*}{Acc}\\ 
		   &F1 &F1 &F1 & & &F1 &F1 &F1 & & \\
	\hline
	CFN-ESA-Emo &- &- &- &66.70 &67.85 &- &- &- &71.04 &70.78 \\
    CFN-ESA-Sent &78.71 &67.06 &70.42 &73.74 &73.75 &88.03 &70.06 &90.99 &84.49 &84.78 \\
	\hline
    \end{tabular}
    CFN-ESA-Emo and CFN-ESA-Sent denote the tasks of emotion classification and sentiment classification, respectively.
    \end{threeparttable}
    \label{tab:sentiment}
\end{table*}

\subsection{Case Study}
\begin{figure}[htbp]
    \centering
    \includegraphics[width=3.4in]{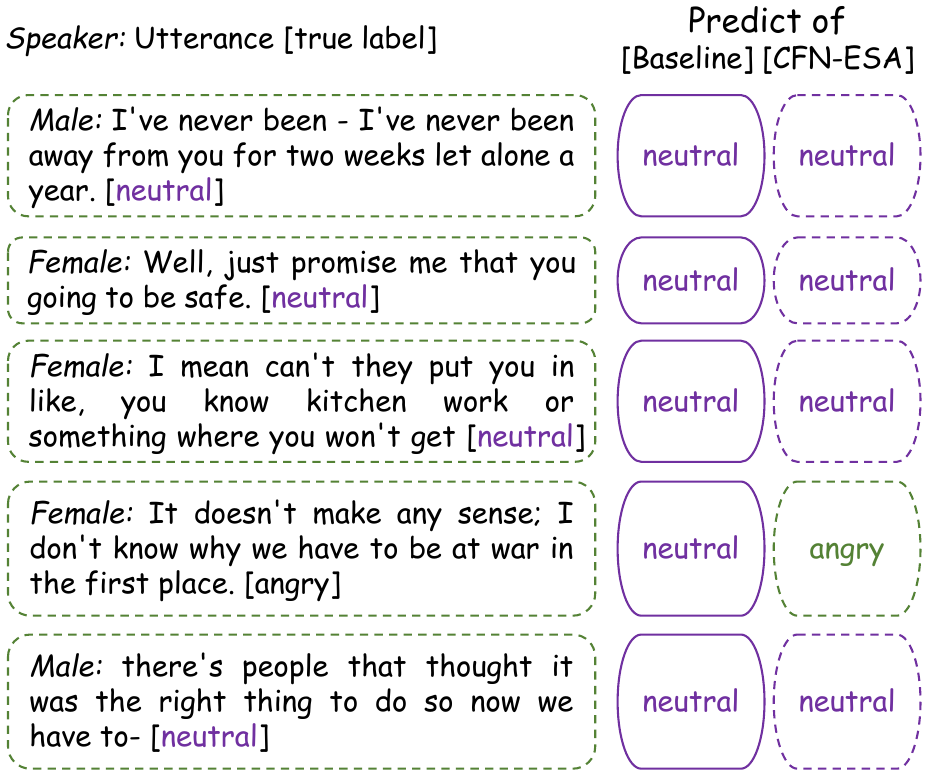}
    \caption{A conversational case in the IEMOCAP dataset.}
    \label{fig:case}
\end{figure}
We discuss a case of emotion shift in this subsection. Fig.~\ref{fig:case} shows a conversational scenario in the IEMOCAP dataset. When a speaker utters several consecutive times with the true emotion \textit{neutral}, most models such as MM-DFN tend to predict the emotion of next utterance as \textit{neutral}. This is due to the fact that these models tend to model based on context, which leads to overly focusing on the contextual information and ignoring the inter-modal self-information. On the contrary, since CFN-ESA can capture emotion-shift information exploiting LESM, which enables the model to strike a trade-off between contextual modeling and self-modeling, e.g., capturing more inter-modal self-information (AKA multimodal complementary information), it identifies the next utterance as the correct emotion \textit{anger}.

\subsection{Error Studies}
\begin{figure*}[htbp]
    \centering
	\subfloat[Confusion Matrix on MELD]{\includegraphics[width=3.0in]{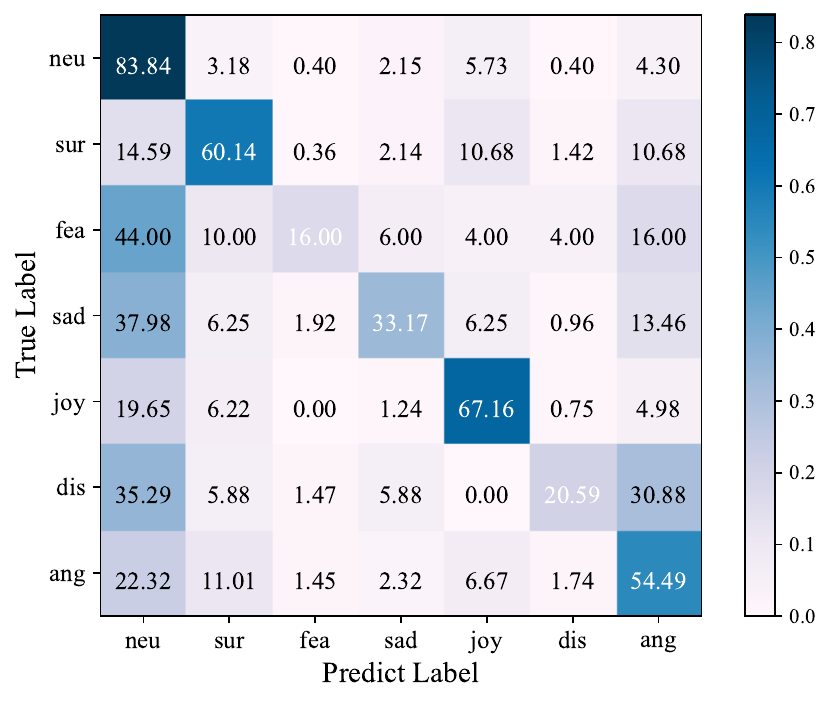}%
    \label{fig:meld_cm}}
    \hfil
    \subfloat[Confusion Matrix on IEMOCAP]{\includegraphics[width=3.0in]{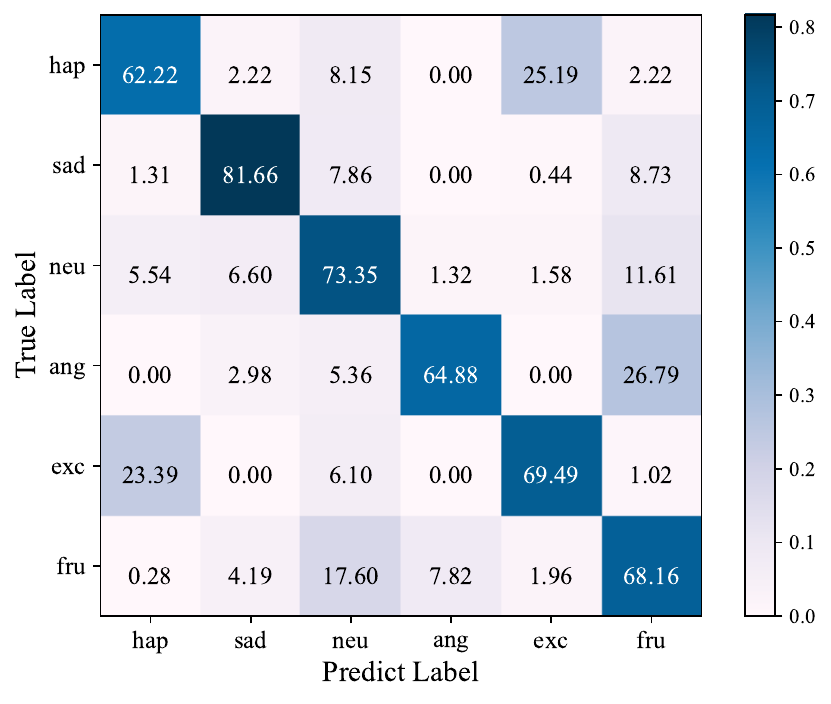}%
    \label{fig:iemocap_cm}}
    \caption{Confusion matrices on the MELD and IEMOCAP datasets. Note that in the confusion matrices, we convert predicted quantities into proportions.}
    \label{fig:cm}
\end{figure*}

Fig.~\ref{fig:cm} shows confusion matrices of our CFN-ESA on the MELD and IEMOCAP datasets. Comparing these two subfigures, it can be concluded that the classification effect of CF-ESA on the IEMOCAP dataset is better than that on the MELD dataset. One primary reason is that MELD is a severely class-imbalanced dataset, where \textit{fear}, \textit{sadness}, and \textit{disgust} belong to the extreme minority classes. As can be witnessed in Fig.~\ref{fig:meld_cm}, the above three classes perform the worst. In most cases, the model tends to recognize them as the majority class (i.e., \textit{neutral}) on the MELD dataset.

Another limitation is that, like most ERC models, our CFN-ESA suffers from the similar-emotion problem. In other words, because the characteristics of some emotions is close to or belongs to the same sentiment, it is difficult for CFN-ESA to differentiate them. For example, on the MELD dataset, the true emotion \textit{disgust} is easily classified as \textit{anger}; on the IEMOCAP data, the proposed CFN-ESA recognizes the true emotion \textit{happy} as \textit{excited} in some cases, as well as detects the true \textit{angry} as \textit{frustrated}. In the case of class imbalance, a minority class itself is hard to recognize correctly, and it is recognized as either the majority class or similar emotion. For example, in Fig.~\ref{fig:meld_cm}, \textit{disgust} is easily categorized as either the majority class \textit{neutral} or similar emotion \textit{anger}. Thus, the similar-emotion problem becomes more severe in class-imbalanced case.

\section{Conclusion}\label{conclusion}
Previous multimodal ERC models exist some flaws, such as (1) failure to distinguish the amount of emotional information in each modality, which causes difficulty in adequately modeling multimodal data; and (2) failure to consider emotion-shift information and overfocusing on capturing intra-modal contextual information, which results in the model not being ability to correctly identify emotions under some emotion-shift scenarios. To address the above issues, we propose a multimodal conversational emotion recognition network, CFN-ESA, to efficiently capture multimodal emotional information, providing a new modeling scheme for the ERC task. Our CFN-ESA mainly contains recurrence based uni-modality encoder (RUME), attention based cross-modality encoder (ACME), and label based emotion-shift module (LESM). The function of RUME is to capture intra-modal contextual information at the conversation level and to narrow the differences in the distribution of multimodal data; ACME takes textual modality as the main source of emotional information, which can effectively extract inter-modal complementary information; and LESM is used to extract emotion-shift information, which guides the main task to reduce intra-modal contextual modeling under emotion-shift scenario, thereby optimizing the emotional expression of the utterance. To demonstrate the effectiveness of CFN-ESA, we conduct comparison experiments and ablation studies on two conversational emotion datasets (i.e., MELD and IEMOCAP). The results of comparison experiments prove that the proposed CFN-ESA outperforms all baselines; the results of ablation studies verify that each component in CFN-ESA can effectively upgrade the performance of the model.

Theoretically, the visual information plays an instrumental role in providing direct emotional cues for the model. Since the visual data often involves a lot of noise from complex environmental scenes, our approach, like most models, has difficulty capturing visual emotional information. Exploring methods that fully utilize the visual modality is a worthwhile research direction in future work. The architecture based on emotion shift merits deeper investigation. The phenomenon of emotion shift is pervasive in dialogue systems and often exerts a detrimental effect on the performance of the model. Consequently, in future work, it is plausible to incorporate emotion-shift prediction as an auxiliary task with the aim to enhance its precision, thereby potentially leading to further improvements in the performance. Also, verifying the generalizability of ERC models is an intriguing subject. For instance, (1) training the model on an independent dataset and subsequently testing its performance on another, thereby providing empirical evidence for its cross-dataset recognition ability; and (2) applying the model to a more challenging real-world dataset in order to substantiate its robustness and practical effectiveness under extreme or unpredictable conditions.

\bibliographystyle{IEEEtran}
\bibliography{IEEEabrv,cfnesa.bib}

\end{document}